\documentclass[conference]{IEEEtran}
\IEEEoverridecommandlockouts
\usepackage{cite}
\usepackage{graphicx}
\usepackage{caption}
\usepackage{bm}  
\usepackage{booktabs}
\usepackage{amsmath} 
\usepackage{array}  
\usepackage{multirow}
\usepackage{stfloats}
\usepackage{algpseudocode}
\usepackage{float}
\usepackage{subcaption}
\usepackage{amsfonts}
\usepackage{stackengine}
\usepackage{makecell}
\usepackage{url}
\usepackage{extdash}
\usepackage[outdir=./]{epstopdf}

\begin{document}

\title{\LARGE \bf Open-Source Reinforcement Learning Environments Implemented in MuJoCo with Franka Manipulator}

\author{Zichun Xu, Yuntao Li, Xiaohang Yang, Zhiyuan Zhao, Lei Zhuang, and Jingdong Zhao\textsuperscript{*}

\thanks{All authors are with the State Key Laboratory of Robotics and Systems, Harbin Institute of Technology, Harbin 150001, Heilongjiang Province, China. \textsuperscript{*}Corresponding author: zhaojingdong@hit.edu.cn}%
}

\maketitle

\begin{abstract}
This paper presents three open-source reinforcement learning environments developed on the MuJoCo physics engine with the Franka Emika Panda arm in MuJoCo Menagerie. Three representative tasks, push, slide, and pick-and-place, are implemented through the Gymnasium Robotics API, which inherits from the core of Gymnasium. Both the sparse binary and dense rewards are supported, and the observation space contains the keys of desired and achieved goals to follow the Multi-Goal Reinforcement Learning framework. Three different off-policy algorithms are used to validate the simulation attributes to ensure the fidelity of all tasks, and benchmark results are also given. Each environment and task are defined in a clean way, and the main parameters for modifying the environment are preserved to reflect the main difference. The repository, including all environments, is available at \url{https://github.com/zichunxx/panda_mujoco_gym}.
\end{abstract}

\begin{IEEEkeywords}
	Reinforcement Learning, MuJoCo, MuJoCo Menagerie
\end{IEEEkeywords}

\section{INTRODUCTION}
Reinforcement learning (RL) has been widely applied to sophisticated decision-making tasks, such as assembly tasks \cite{yamada2020motion} and connector insertion \cite{zhao2023learning}, and has performed well even with high-dimensional vision input tensors \cite{zhang2021sim2real}. This advantage is due to the fact that the goal and reward can be implicitly defined, and the agent can explore the environment to gather valuable actions. In recent years, many contributions have been made to the RL community, especially Gymnasium \cite{towers_gymnasium_2023}, which is a fork of the OpenAI Gym library and provides lots of APIs and wrappers to formalize the standard pipeline.

Nevertheless, both the contact-rich task \cite{zakka2023robopianist} and reward engineering \cite{singh2019end}, which dominate the training process, preclude training in the open world to some extent. Therefore, a realistic simulation environment can further minimize the sim-to-real gap and serve as a reference for deployment in the real scenario. So far, convincing researches \cite{pinto2018asymmetric, chebotar2019closing, tang2023industreal} have demonstrated that training in simulation is informative and the derived policy can be transferred to real scenarios, which also raises high demands on the computational speed and accuracy of the physics engine. Some options can be considered to realize parallel simulation, e.g., MuJoCo\footnote[1]{https://mujoco.org/}, Bullet\footnote[2]{https://pybullet.org/}, and PhysX\footnote[3]{https://github.com/NVIDIAGameWorks/PhysX}, and the accessible environments built within them lower the entry barrier for RL researchers.

\begin{figure}[t]
	\centering
	\includegraphics[scale=.11]{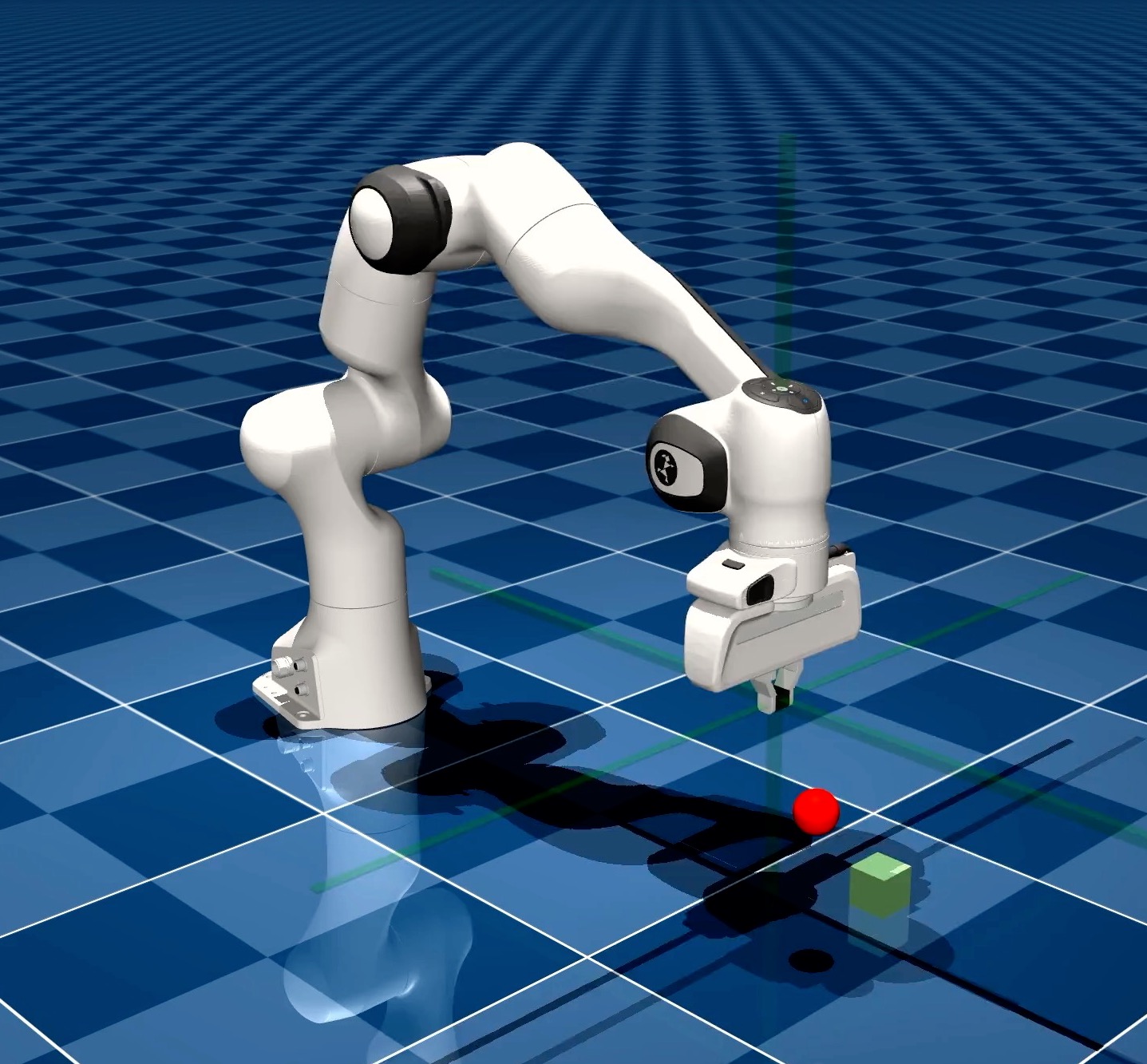}
	\caption{Base simulation environment implemented with the Franka model in MuJoCo Menagerie.}
	\label{intro}
\end{figure}

MuJoCo enables different types of contact dynamics and actuation models with various optimization solvers, which makes it a popular choice and has been adopted in many studies \cite{lee2021ikea,nair2018overcoming}. Meanwhile, a collection of fine-tuned robot models is published in MuJoCo Menagerie\footnote[4]{\url{http://github.com/google-deepmind/mujoco_menagerie}}. Gymnasium \cite{towers_gymnasium_2023} integrates with many out-of-box classic RL environments developed on MuJoCo, such as \texttt{Humanoid} and \texttt{Ant}. With the development of operation tasks, environments focused on collaborative manipulators are gradually receiving more attention. Gymnasium Robotics\footnote[5]{\url{http://github.com/Farama-Foundation/Gymnasium-Robotics}} is presented including the Fetch \cite{plappert2018multi} and Franka Kitchen \cite{gupta2019relay} environments, in which the Fetch environments consist of four representative tasks: \texttt{FetchReach}, \texttt{FetchPush}, \texttt{FetchSlide}, and \texttt{FetchPickAndPlace}. These continuous control tasks are provided as guidance on how to define similar behaviors and the goal-conditioned observation space. Furthermore, Robosuite \cite{robosuite2020}, which is a modular simulation framework powered by MuJoCo, encapsulates some benchmark manipulation environments with seven commercially available arms. 

Pybullet\footnote[6]{\url{https://github.com/bulletphysics/bullet3}}, which is a python module for robotics simulation, also encapsulates a suite of basic Gym environments. Beyond that, the KUKA IIWA arm is incorporated to construct \texttt{KukaBulletEnv} and \texttt{KukaCamBulletEnv}, in which the observation for the latter is camera pixels. Inspired by the Fetch environments, panda-gym \cite{gallouedec2021panda} is developed on Pybullet with two extra tasks, \texttt{PandaFlip} and \texttt{PandaStack}. All environments in panda-gym follow the Multi-Goal RL framework and are defined as the robot and task in an individual way for more flexible development.

Isaac Gym \cite{makoviychuk2021isaac}, built on NVIDIA PhysX, features accelerating the training process on GPU and also incorporates a host of robotics-related environments. Factory \cite{narang2022factory} resorts to reproduce the realistic contact-rich assembly scenario in simulator, and thus three environments are developed to solve the assembly tasks on the NIST task board, e.g., \texttt{FrankaNutBoltEnv}, \texttt{FrankaInsertionEnv}, and \texttt{FrankaGearsEnv}. Single and dual KUKA arm manipulation environments with each of the three task variants (reorientation, regrasping, and grasp-and-throw) are also created with the 16-DoF Allegro Hand \cite{petrenko2023dexpbt}.

Motivated by the preceding work, we aim to provide a unified framework that encompasses benchmark environments powered by MuJoCo for two key reasons: First, compared to other engines, MuJoCo is designed for robotics and performs impressively with the fastest and most accurate simulations \cite{erez2015simulation}. Some benchmark environments built on MuJoCo are easy to build and can be simply deployed with only a single CPU. NVIDIA's Isaac Sim requires a high-performance GPU as the hardware foundation for building environments, which is a disincentive for beginners. Second, the modular design of some environments, like Robosuite, is user- and developer-friendly, but it is not intuitive for researchers to build their own environments on top of it and tune simulation options. Thus, from this perspective, building environments separately in a clean way is preferable.

\begin{figure*}[th]
	\centering
	\includegraphics[scale=.4]{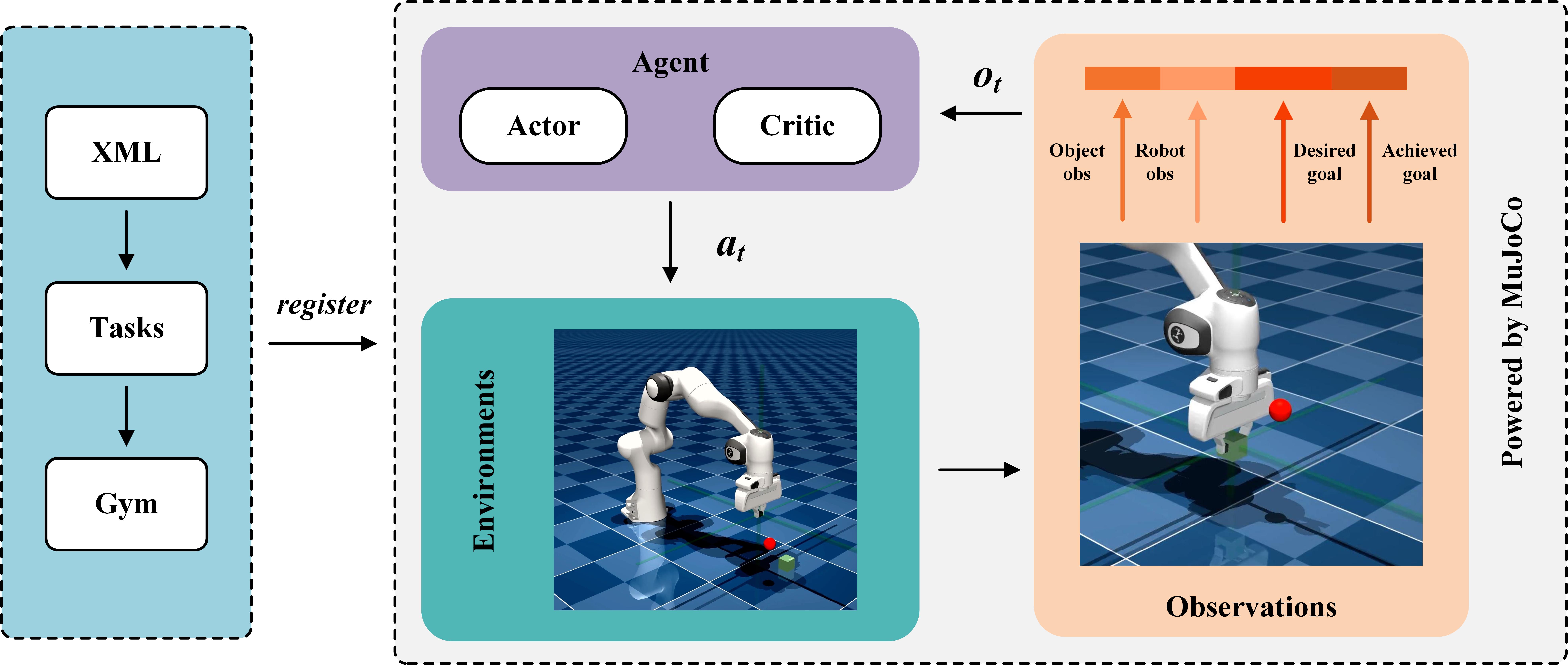}
	\caption{Overview of the benchmark environment, including registration and the training process.}
	\label{flow}
\end{figure*}

Toward this end, some manipulation tasks performed by the Franka arm are built with necessary dependencies and simplified modules. The simulation model of the Franka arm is obtained in MuJoCo Menagerie. The reason for this is that MuJoCo Menagerie offers a wider choice of high-quality robot models, which do not only cover manipulators, and few benchmark environments are built on them. This work can be treated as an extension of panda-gym \cite{gallouedec2021panda} and migrated from Bullet to MuJoCo. The benchmark results with Deep Deterministic Policy Gradient (DDPG) \cite{lillicrap2015continuous}, Soft Actor Critic (SAC) \cite{haarnoja2018softa,haarnoja2018soft}, and Truncated Quantile Critics (TQC) \cite{kuznetsov2020controlling} are also provided. In summary, the main contributions of this paper are twofold:

\begin{itemize}
	\item[1)] Three benchmark environments are constructed based on MuJoCo Menagerie and powered by MuJoCo.
	\item[2)] DDPG, SAC, and TQC are trained over all tasks, and benchmark results are provided with essential analyses.
\end{itemize}

The remainder of this paper is organized as follows: Sec. \ref{sec_pre} first reviews all the off-policy RL algorithms used in this paper. Sec. \ref{envs} introduces the implementation details for all environments, and Sec. \ref{simu} presents the benchmarking results over random seeds with essential analysis. Finally, Sec. \ref{cons} concludes this paper and illustrates some future works.

\section{PRELIMINARIES}\label{sec_pre}
Each task addressed in this paper can be formulated as a finite-horizon Markov decision process consisting of state space $\mathcal{S}$, action space $\mathcal{A}$, starting state distribution $\rho_0$, transition probability function $\mathcal{P}: \mathcal{S} \times \mathcal{A} \to \mathcal{S}$, reward function $r: \mathcal{S} \times \mathcal{P} \to \mathcal{R}$, and discount factor $\gamma \in [0,1]$. After resetting the environment, each episode will start with $s_0 \in \rho_0$. The policy $\pi(a_t|s_t)$ is the mapping from $\mathcal{S}$ to $\mathcal{A}$, from which the agent will take the action $a_t\in\mathcal{A}$ at the timestep $t$. Then the reward $r_t$, the new state $s_{t+1}\sim\mathcal{P}$, and the signal $d$ are received at the next timestep, where $d$ is used to distinguish whether the current episode is terminated or truncated. The above transition tuple $(s_t, a_t, r_t, s_{t+1}, d)$ will be stored in replay buffer $\mathcal{D}$ for training. The goal for RL is to select an optimal policy $\pi^*$ to maximize the expected return $\mathbb{E}[\sum_{i=t}^{T}\gamma^{i-t}r_i]$, where $T$ denotes the episode horizon.
\subsection{Deep Deterministic Policy Gradient}
DDPG is an off-policy algorithm that concurrently learns a deterministic policy $\pi_\theta$ and a Q-function $Q_\phi(s, a)$ for continuous action, where $\theta$ and $\phi$ are initial policy and Q-function network parameters. At each timestep, the policy and Q-function networks are updated by sampling a minibatch $B=\{(s_t, a_t, r_t, s_{t+1}, d)\}$ from $\mathcal{D}$. Notably, $a_t$ is obtained by adding the mean-zero Gaussian noise at the training time and saved in $\mathcal{D}$. After that, $Q_\phi(s_t, a_t)$ is updated utilizing gradient descent with 
\begin{equation}
	\nabla_\phi \frac{1}{|B|} \sum_{\left(s_t, a_t, r_t, s_{t+1}, d\right) \in B}\biggl(Q_\phi(s_t, a_t)-y\left(r, s_{t+1}, d\right)\biggr)^2,
	\label{ddpg_1}
\end{equation}
where the target is calculated by 
\begin{equation}
	y(r,s_{t+1},d)=r+\gamma(1-d)Q_{\phi_{\rm{targ}}}\biggl(s_{t+1},\pi_{\theta_{\rm{targ}}}(s_{t+1})\biggr). 
\end{equation}
$\pi_\theta$ is updated using 
\begin{equation}
	\nabla_\phi \frac{1}{|B|} \sum_{s_t \in B}Q_\phi\left(s_t,\pi_\theta(s_t)\right)
\end{equation}
at a certain frequency. The weights of target networks $\theta_{\rm{targ}}$ and $\phi_{\rm{targ}}$ will be updated by polyak averaging with a certain step delay. 
\begin{figure*}[th]
	\centering
	\captionsetup[subfigure]{skip=6pt,slc=off,margin={40pt, 0pt},labelfont=normalfont}
	\hspace{0.4cm}
	\subcaptionbox{FrankaPush\label{push_scene}}[5cm][c]{\includegraphics[scale=0.06]{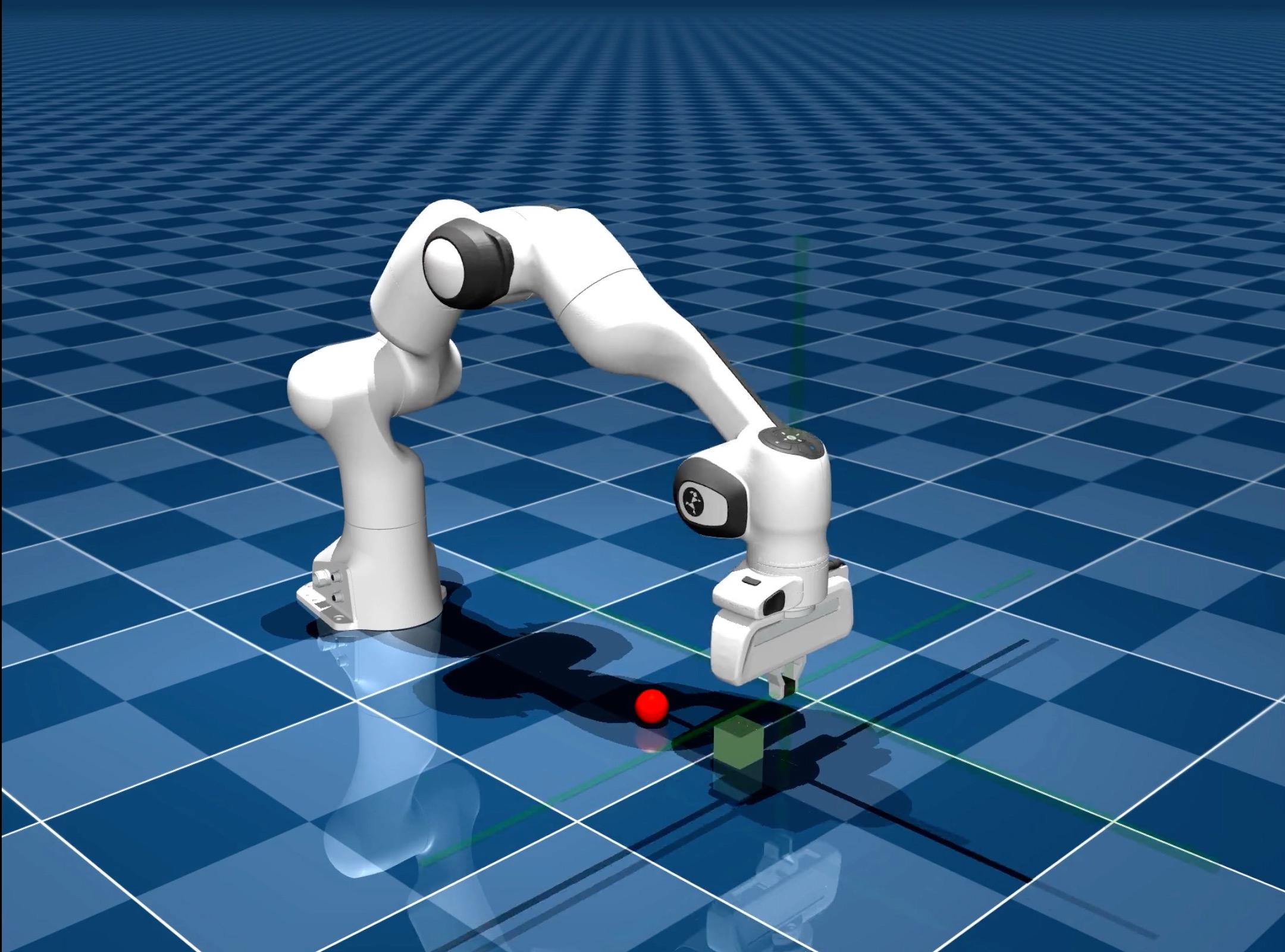}}
	\hspace{0.4cm}
	\subcaptionbox{FrankaSlide\label{slide_scene}}[5cm][c]{\includegraphics[scale=0.06]{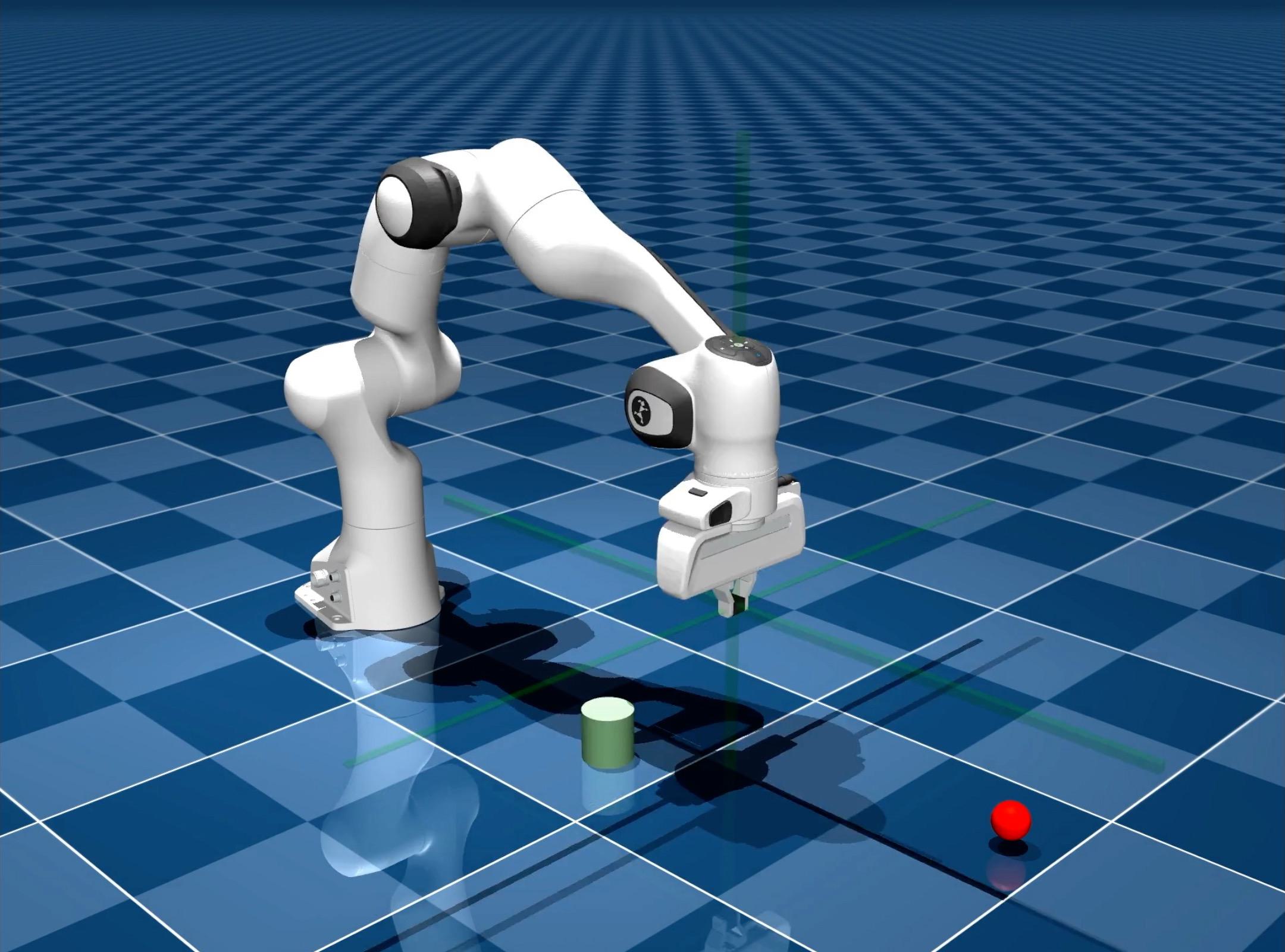}}
	\hspace{0.4cm}
	\captionsetup[subfigure]{skip=6pt,slc=off,margin={25pt, 0pt},labelfont=normalfont}
	\subcaptionbox{FrankaPickAndPlace\label{pick_scene}}[5cm][c]{\includegraphics[scale=0.06]{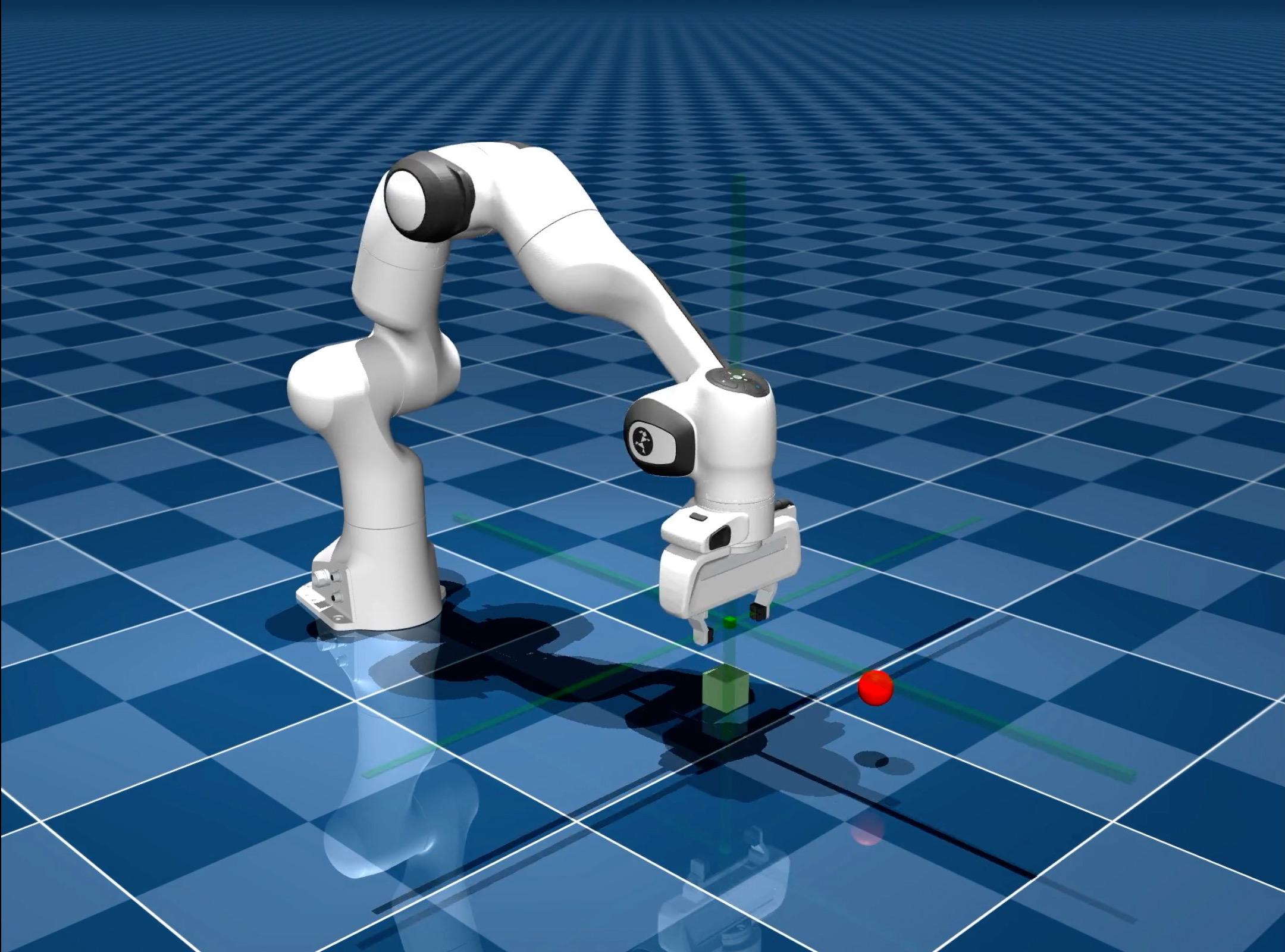}}
	\caption{Three proposed environments with the Franka model in MuJoCo Menagerie, in which the red point indicates the target.}
	\vspace{0pt}
	\label{scenes}
\end{figure*}
\subsection{Soft Actor Critic}
In contrast to DDPG, SAC trains a stochastic policy and two Q-functions $Q_{\phi_1}$ and $Q_{\phi_2}$ based on the entropy framework. For the variant with the fixed entropy temperature coefficient $\alpha$, the target in Eq.~\ref{ddpg_1} is given by
\begin{equation}
	\begin{aligned}[b]
		y(r,s_{t+1},d)=r+\gamma(1-d)\biggl(\min_{j=1,2}Q_{\phi_{{\rm{targ}}, j}}(s_{t+1},\tilde{a}_{t+1}) \\
		-\alpha\log\pi_{\theta}(\tilde{a}_{t+1}|s_{t+1})\biggr),
	\end{aligned}
\end{equation}
where ${\tilde{a}}_{t+1}\sim\pi_{\theta}(\cdot|s_{t+1})$ is sampled from the current policy via the reparameterization trick. The policy is updated by gradient ascent with
\begin{equation}
	\nabla_\phi \frac{1}{|B|} \sum_{s_t \in B}\biggl(\min_{i=1,2}Q_{\phi_i}\left(s_t,\tilde{a}_{t+1}\right)-\alpha\log\pi_{\theta}(\tilde{a}_{t+1}|s_t)\biggr),
\end{equation}
which uses the minimum of $Q_{\phi_1}$ and $Q_{\phi_2}$. In addition, $\alpha$ can be dynamically adjusted concerning the loss
\begin{equation}
	\frac{1}{|B|} \sum_{s_t \in B}\biggl(-\log\alpha\cdot\bigl(\log\pi_{\theta}(\tilde{a}_{t+1}|s_t)+\mathcal{H}_T\bigr)\biggr)
\end{equation}
and thus balances exploration and exploitation, where $\mathcal{H}_T=-\dim\mathcal{A}$.
\subsection{Truncated Quantile Critics}
TQC is an extension of SAC but learns the distributional representation for $N$ Q-function critics. Each critic $Q_{\phi_n}(s_t,a_t), n \in \{1,...,N\}$ estimates the distribution over the sum of discount rewards and contains $M$ atoms, i.e., 
\begin{equation}
	Q_{\phi_n}(s_t,a_t)=\frac{1}{M}\sum_{m=1}^M\delta(\theta_{\phi_n}^m(s_t,a_t)). 
\end{equation}
All atoms of $Q_{\phi_{{\rm{targ}},n}}(s_{t+1},\tilde{a}_{t+1})$ are pooled into a set 
\begin{equation}
	\begin{aligned}[b]
		\mathcal{Z}(s_{t+1},\tilde{a}_{t+1}) := \Bigl\{\theta_{\phi_{{\rm{targ}},n}}^m(s_{t+1},\tilde{a}_{t+1})|n \in [1...N],\\
		m \in [1...M]\Bigr\}
	\end{aligned}
\end{equation}
and sorted in ascending order $z_i(s_{t+1}, \tilde{a}_{t+1}), i \in \{1...NM\}$, in which the $j \times N$ largest elements are truncated. The remaining $k \times N$ atoms form a target distribution
\begin{equation}
		Y(s_t, a_t) := \frac{1}{kN}\sum_{i=1}^{kN}\delta({y_i}(r, s_{t+1}, d)), k = M-j
\end{equation}
that is used to calculate the critic loss, where  
\begin{equation}
	\begin{aligned}[b]
		y_i(r, s_{t+1}, d) = & r + \gamma(1-d)\biggl(z_i(s_{t+1}, \tilde{a}_{t+1}) \\
		&-\alpha \log\pi_\theta (\tilde{a}_{t+1}|s_{t+1})\biggr), i \in \{1...kN\}.
	\end{aligned}
\end{equation}
The critic loss can be derived by
\begin{equation}
	\begin{aligned}[b]
	    \cfrac{1}{kNM}\sum_{m=1}^{M}\sum_{i=1}^{kN}\rho_{\tau_m}^H\biggl(y_i(r, s_{t+1}, d)-\theta_{\phi_n}^{m}(s_t,a_t)\biggr)
	\end{aligned}
\end{equation}
with the Huber quantile loss
\begin{equation}
	\rho_{\tau_m}^H(u) = |\tau_m - \mathbb{I} (u<0)| \mathcal{L}_H^1(u),
\end{equation}
where $\mathcal{L}_H^1(u)$ is the Huber loss with the linear/quadratic changepoint 1. The policy of TQC is updated as in SAC to minimize
\begin{equation}
	\frac{1}{|B|} \sum_{s_t \in B}\biggl(\alpha\log\pi_{\theta}(\tilde{a}_{t+1}|s_t)-\cfrac{1}{NM}\sum_{m,n=1}^{M,N}\theta_{\phi_n}^{m}(s_t,\tilde{a}_{t+1})\biggr),
\end{equation}
where $\theta_{\phi_n}^{m}(s_t,\tilde{a}_{t+1}), n \in [1...N],m \in [1...M]$ are not truncated.

\section{ENVIRONMENTS}\label{envs}
All tasks in this paper follow the Multi-Goal RL framework, which is similar to panda-gym and Fetch. However, the main difference with panda-gym is that MuJoCo is employed as the physics engine with the Franka model in MuJoCo Menagerie. The settings for each environment, including feasible simulation options and attributes, are given in a separate XML document encapsulating each task. The training script is integrated based on the \texttt{MujocoRobotEnv} API in Gymnasium Robotics. The included tasks can be ranked in the ascending order of complexity:
\begin{itemize}
	\item \texttt{FrankaPush} The object needs to be pushed by the arm to the target on the floor.
	\item \texttt{FrankaSlide} The arm will strike the puck toward the target on the floor, where the floor is slippery with low friction.
	\item \texttt{FrankaPickAndPlace} The arm first picks an object and moves to the desired target, where the target has a certain probability of being chosen above the floor.
\end{itemize}

\begin{table}[t!]
	\vspace{4pt}
	\centering
	\caption{Hyperparameters used for the training of all off-policy algorithms}
	\label{hyper}
	\begin{tabular}{@{}cc@{}}
	\toprule
	Parameter                     & Value                                 \\ 
	\midrule
	Action size 				  & 3 (Push, Slide), 4 (Pick \& Place)    \\
	Observation size 			  & 18 (Push, Slide), 19 (Pick \& Place)  \\
	\multirow{2}{*}{Network size} & $[256,256,256]$ (Pick \& Place, Push) \\
								  & $[512,512,512]$ (Slide)               \\  
	\multirow{2}{*}{Batch size} & $512$ (Pick \& Place, Push) 		  	  \\
								  & $2048$ (Slide)               		  \\  
	Buffer size                   & $10^6$                                \\
	Optimizer 					  & Adam 								  \\
	Action noise           		  & $\mathcal{N}(0,0.2)$ (DDPG, SAC)      \\
	Learning rate                 & $0.001$                               \\
	Polyak update                 & $0.05$                                \\
	Discount factor               & $0.95$                                \\
	Evaluation frequency          & $2,000$                               \\
	Evaluation episode            & $15$                                  \\
	\multirow{2}{*}{Training steps } & $5\times10^5$ (Pick \& Place, Push)  \\
								  & $10^6$ (Slide)                 		    \\
	HER strategy                  & Future                                  \\
	Number of HER per transition  & $4$                       			    \\
	Reward 						  & Sparse                  			    \\ 
	Number of critics        	  & $2$ (TQC only) 							\\
	Number of quantiles			  & $25$ (TQC only)                         \\
	Quantiles to drop per critic 	  & $2$ (TQC only)						\\
	Entropy regularization coefficient & Autotune (SAC, TQC)				\\
	\bottomrule   
	\end{tabular}
\end{table}

The above environments are rendered in Fig.~\ref{scenes} and implemented with the ($7+1$)-DOF (i.e., seven arm joints and one gripper joint) Franka arm with its default parallel gripper. The orientation of the end-effector (EE) is fixed with a grasp pose, and thus the action space is 4-dim for the \texttt{FrankaPickAndPlace} task: the 3-dim movement of EE and the actuation of the gripper. However, the gripper is locked for the \texttt{FrankaPush} and \texttt{FrankaSlide} tasks. A mocap\footnote[1]{\url{https://mujoco.readthedocs.io/en/stable/XMLreference.html#body-mocap}} body is bound to EE to maintain its orientation, and the joint angles are derived depending on its position variation, which is equivalent to the inverse kinematics algorithm. The simulation timestep for MuJoCo is 2 ms based on the speed-accuracy trade-off. It can be further reduced for more stable contacts but with massive computational costs. An extremely small timestep wastes CPU time without meaningful improvement in accuracy. Due to the smaller contact area of the Franka gripper pads, the friction coefficients and contact-related parameters are further tuned, which are given in the XML documents. 

For all tasks in this paper, the observation space includes the Cartesian position and linear velocity of EE ($x_e$, $v_e \in  \mathbb{R}^3$) and the pose (i.e., Cartesian position and Euler angles), linear, and angular velocities of the object ($x_b$, $e_b$, $v_b$, $w_b \in  \mathbb{R}^3$). An additional finger width (1-dim) is added to the observation space for the \texttt{FrankaPickAndPlace} task. The achieved and desired goals, which are keys in the dictionary observation space, are the positions of the object and goal, respectively. Similarly, two variants of reward are retained for all tasks. For the sparse reward, the agent that moves within the target area (i.e., a given distance threshold from the target position) gets a reward of 0 and -1 otherwise. The dense reward is simply defined as a linear function of the distance between the achieved and desired goals.

\begin{table}[t!]
	\vspace{4pt}
	\centering
	\caption{Success rate of each algorithm in different tasks}
	\label{results}
	\begin{tabular}{@{}c c c c}
	\toprule
	Tasks         	 & DDPG   &  SAC    & TQC    \\ 
	\midrule
	Push   			 & 11.7 $\pm$ 0.3  & 100.0 $\pm$ 0.0  & 100.0 $\pm$ 0.0 \\
	Slide   		 & 71.7 $\pm$ 0.5  & 85.0  $\pm$ 0.4  & 85.1  $\pm$ 0.4 \\
	Pick \& Place    & 58.3 $\pm$ 0.4  & 75.0  $\pm$ 0.5  & 100.0 $\pm$ 0.0 \\
	\bottomrule   
	\end{tabular}
\end{table}

\section{EVALUATION}\label{simu}
\begin{figure*}[th]
	\centering
	\captionsetup[subfigure]{skip=2pt,slc=off,margin={42pt, 0pt},labelfont=normalfont}
	\subcaptionbox{FrankaPush\label{push}}[5cm][c]{\includegraphics[scale=0.3]{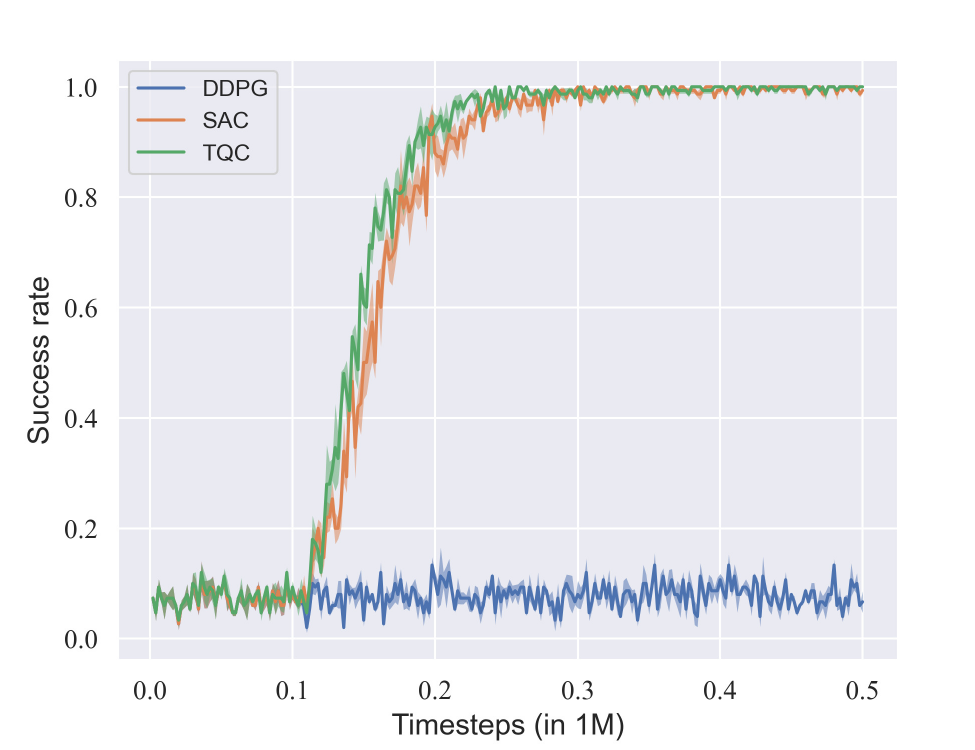}}
	\hspace{0.2cm}
	\subcaptionbox{FrankaSlide\label{slide}}[5cm][c]{\includegraphics[scale=0.3]{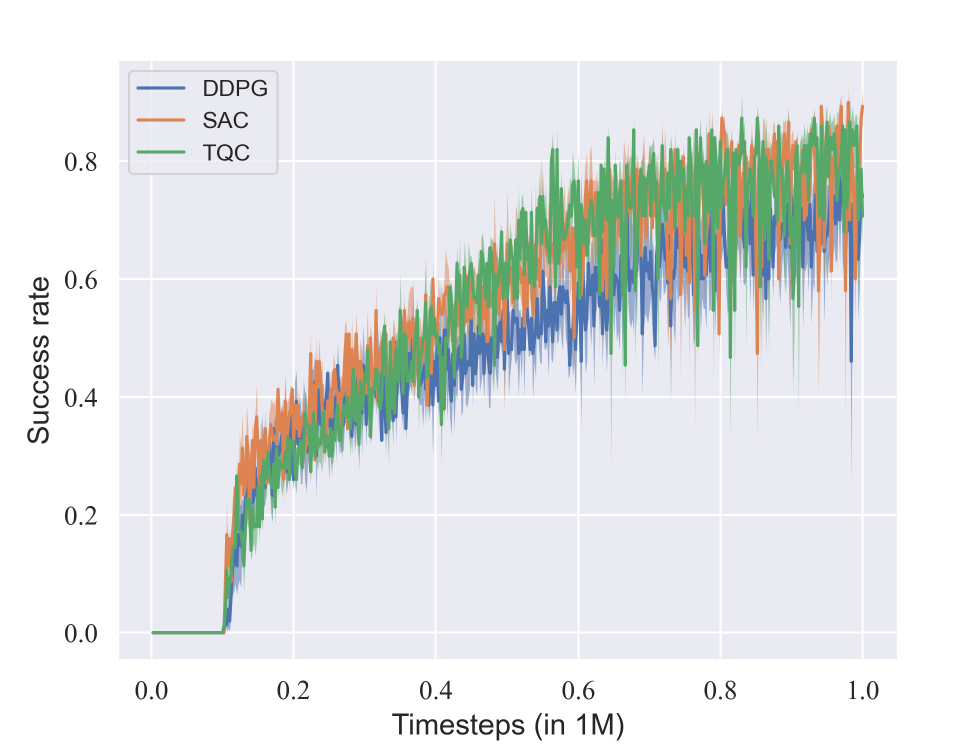}}
	\hspace{0.2cm}
	\captionsetup[subfigure]{skip=2pt,slc=off,margin={28pt, 0pt},labelfont=normalfont}
	\subcaptionbox{FrankaPickAndPlace\label{pick}}[5cm][c]{\includegraphics[scale=0.3]{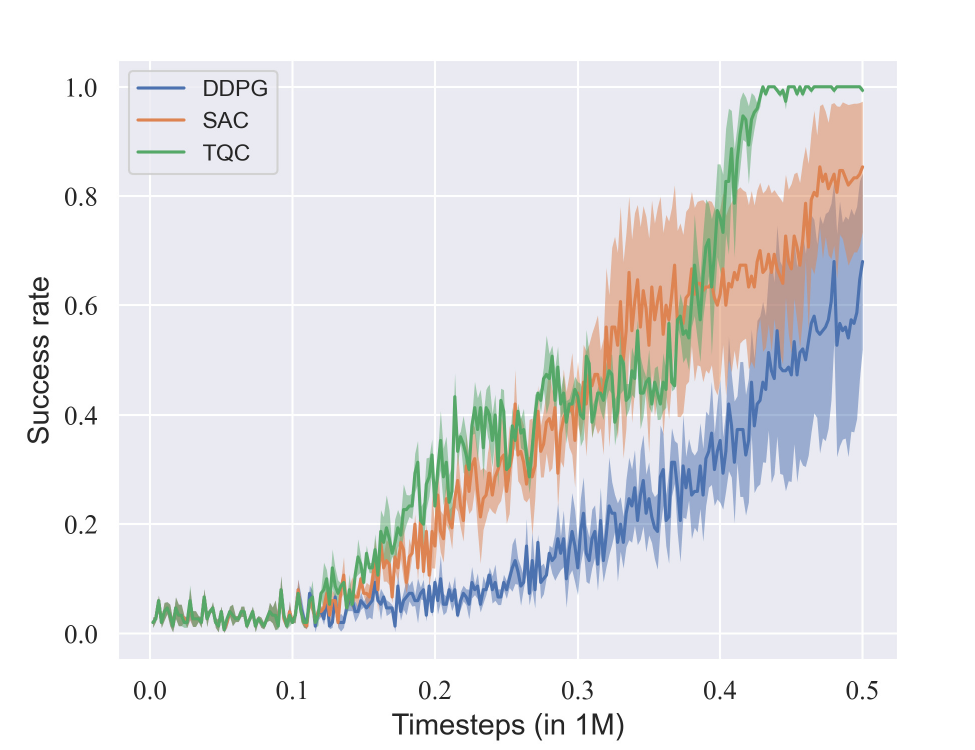}}
	\caption{Median success rates and standard deviations indicated by shaded areas over three random seeds. }
	\label{evaluation}
	\vspace{0pt}
\end{figure*}
\begin{figure*}[t]
	\centering
	\captionsetup[subfigure]{skip=4pt,slc=off,margin={230pt, 0pt},labelfont=normalfont}
	\subcaptionbox{FrankaPush\label{push_seq}}[\textwidth][c]{
		\includegraphics[scale=0.05]{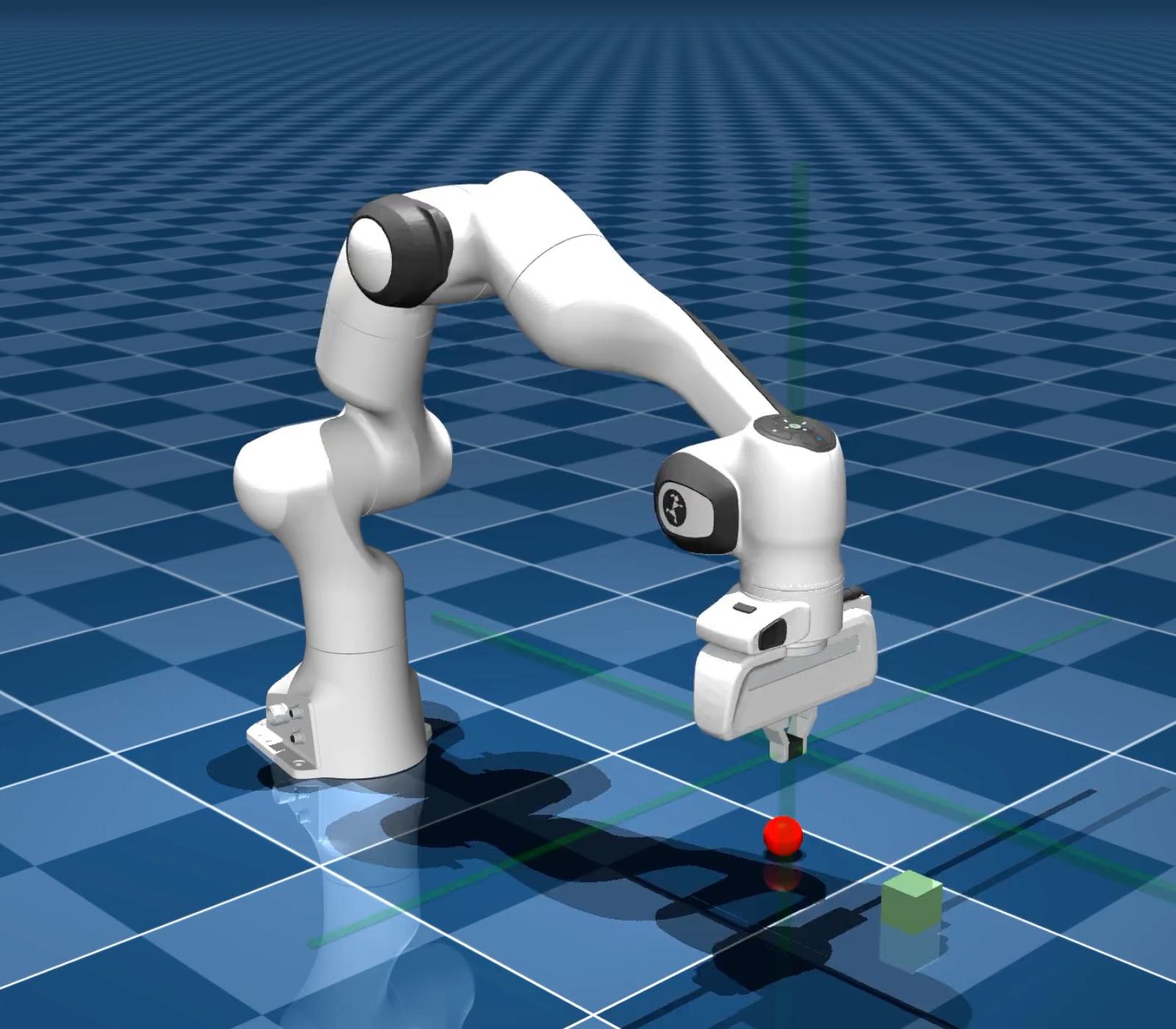}\hspace{0.06cm}
		\includegraphics[scale=0.05]{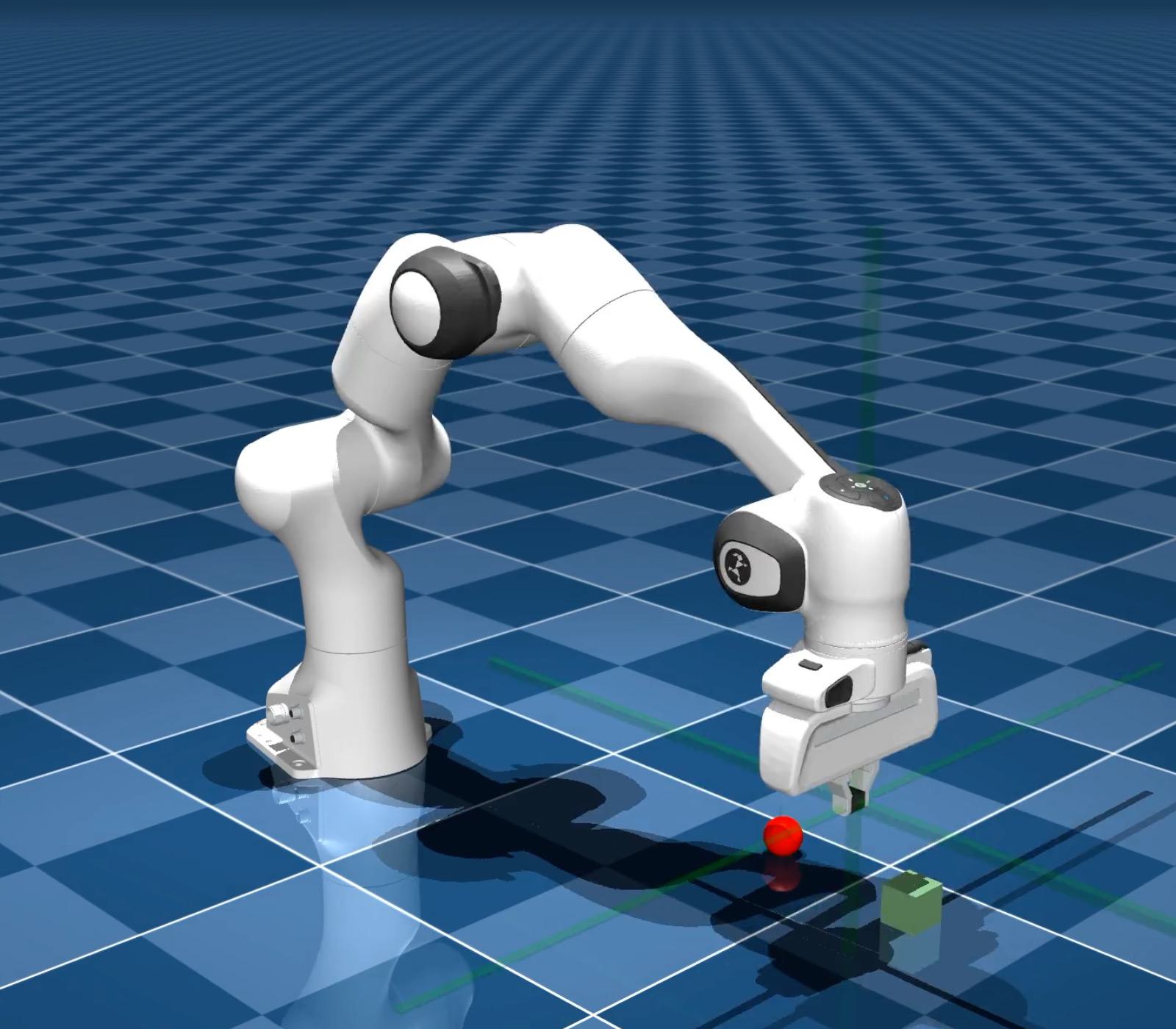}\hspace{0.06cm}
		\includegraphics[scale=0.05]{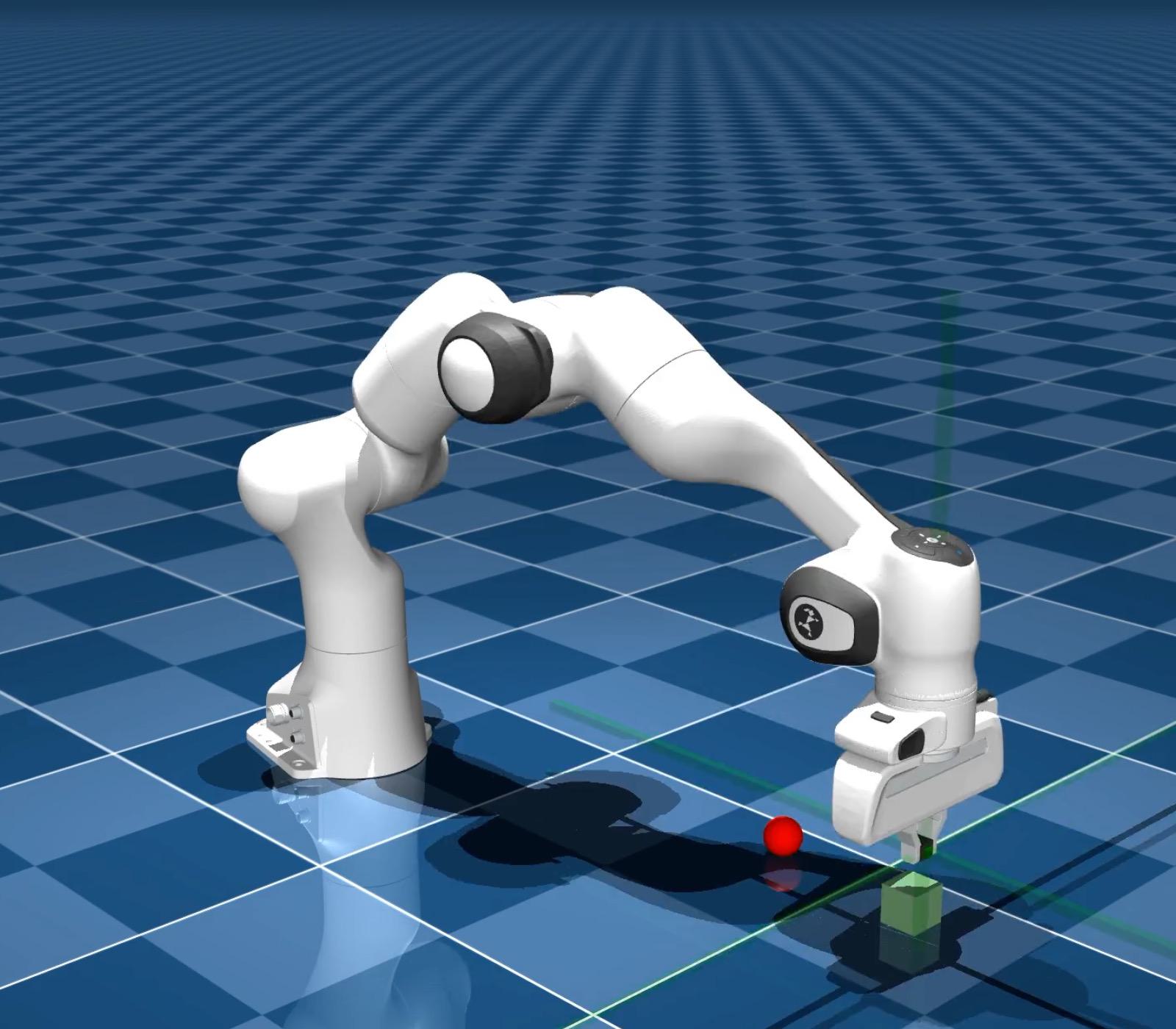}\hspace{0.06cm}
		\includegraphics[scale=0.05]{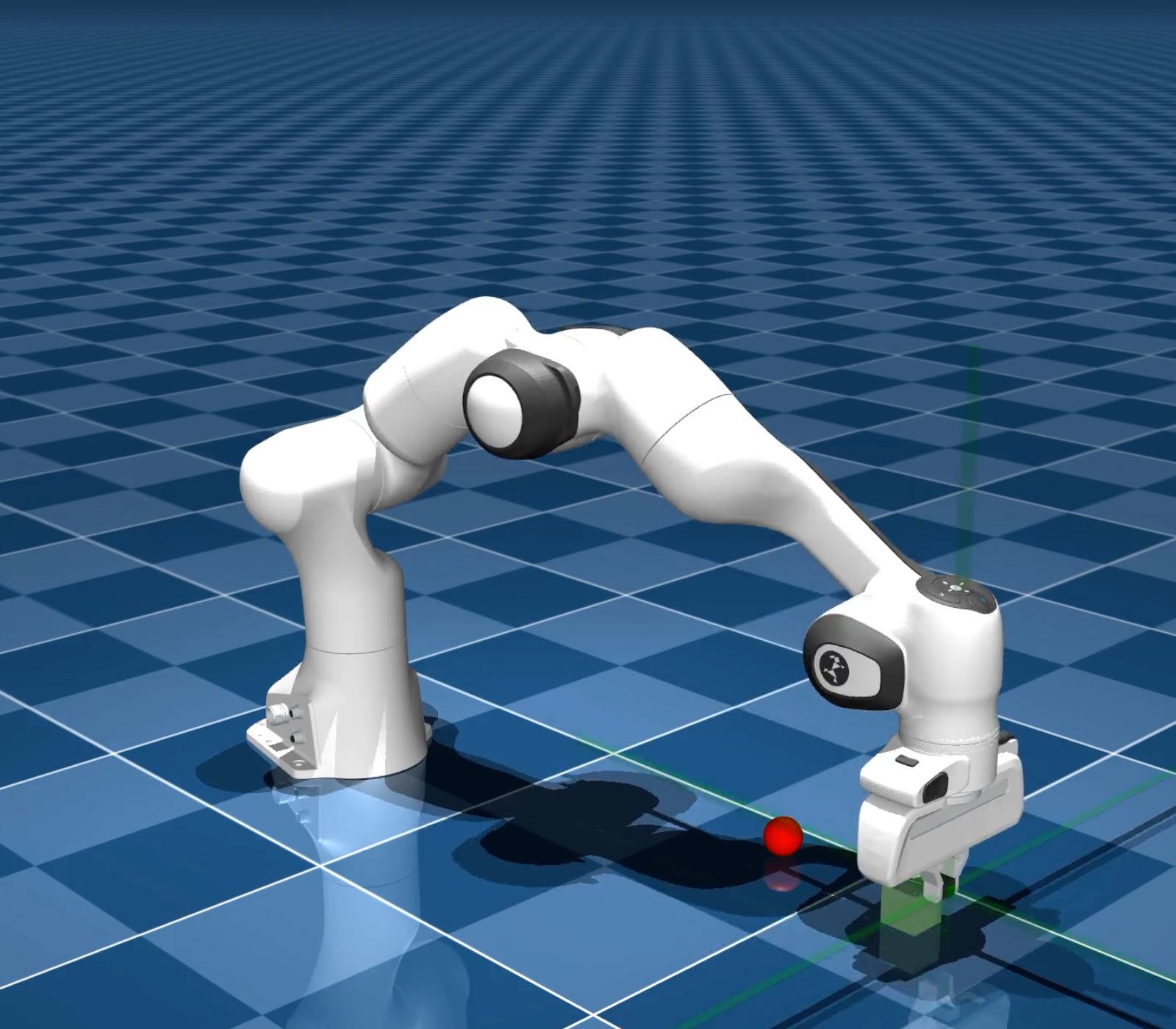}\hspace{0.06cm}
		\includegraphics[scale=0.05]{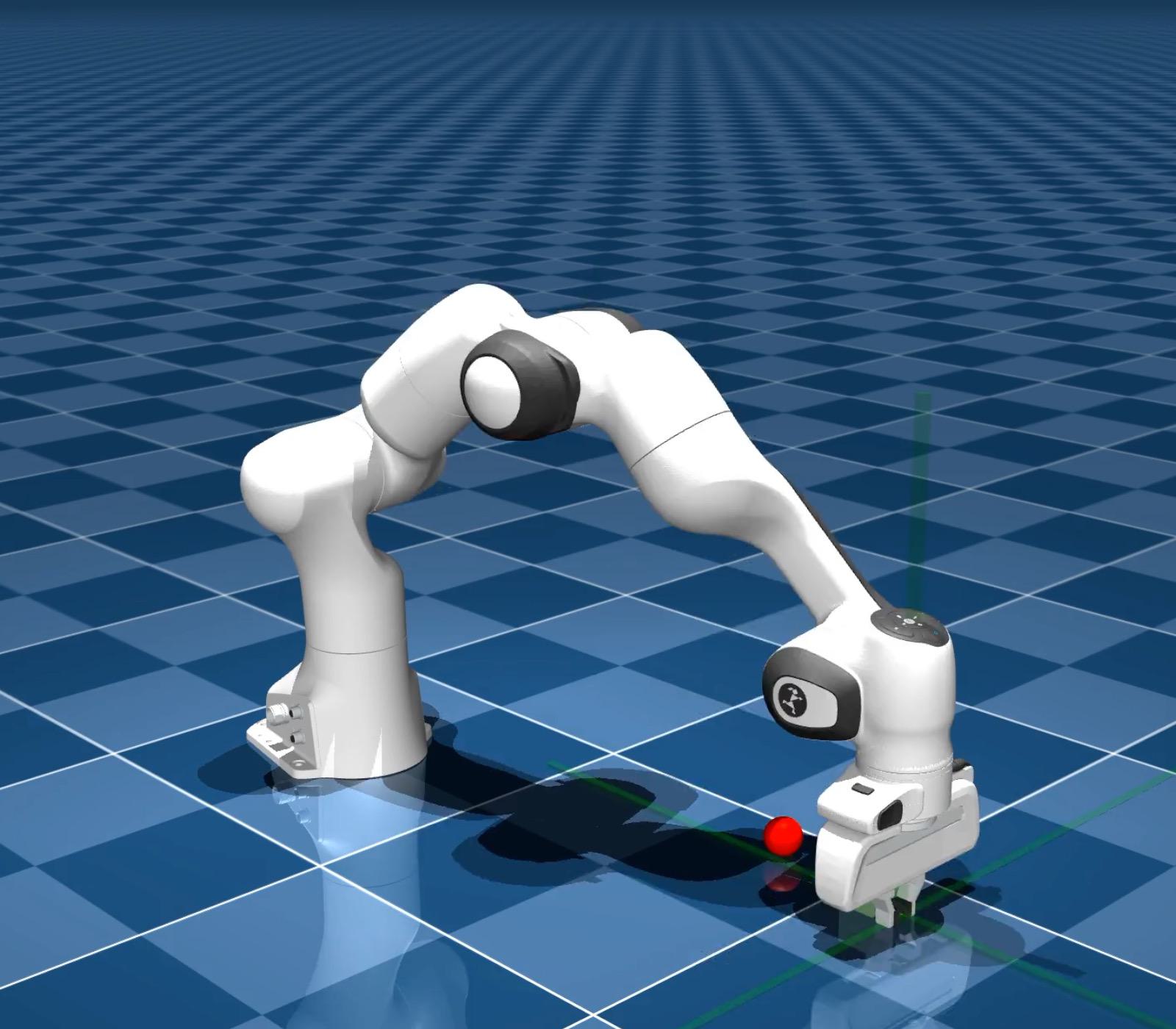}}

	\vspace{0.1cm}
	\subcaptionbox{FrankaSlide\label{slide_seq}}[\textwidth][c]{
		\includegraphics[scale=0.05]{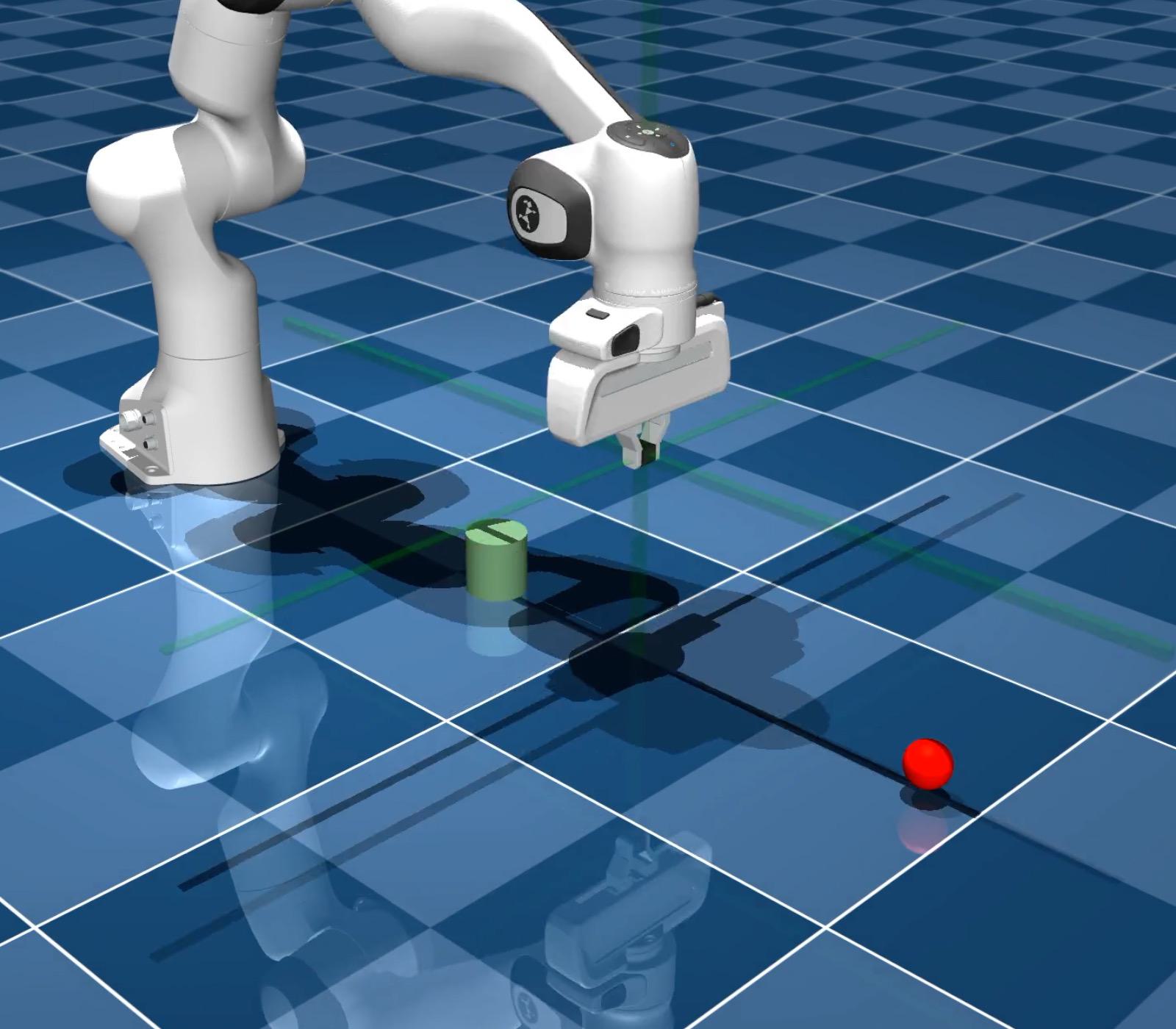}\hspace{0.06cm}
		\includegraphics[scale=0.05]{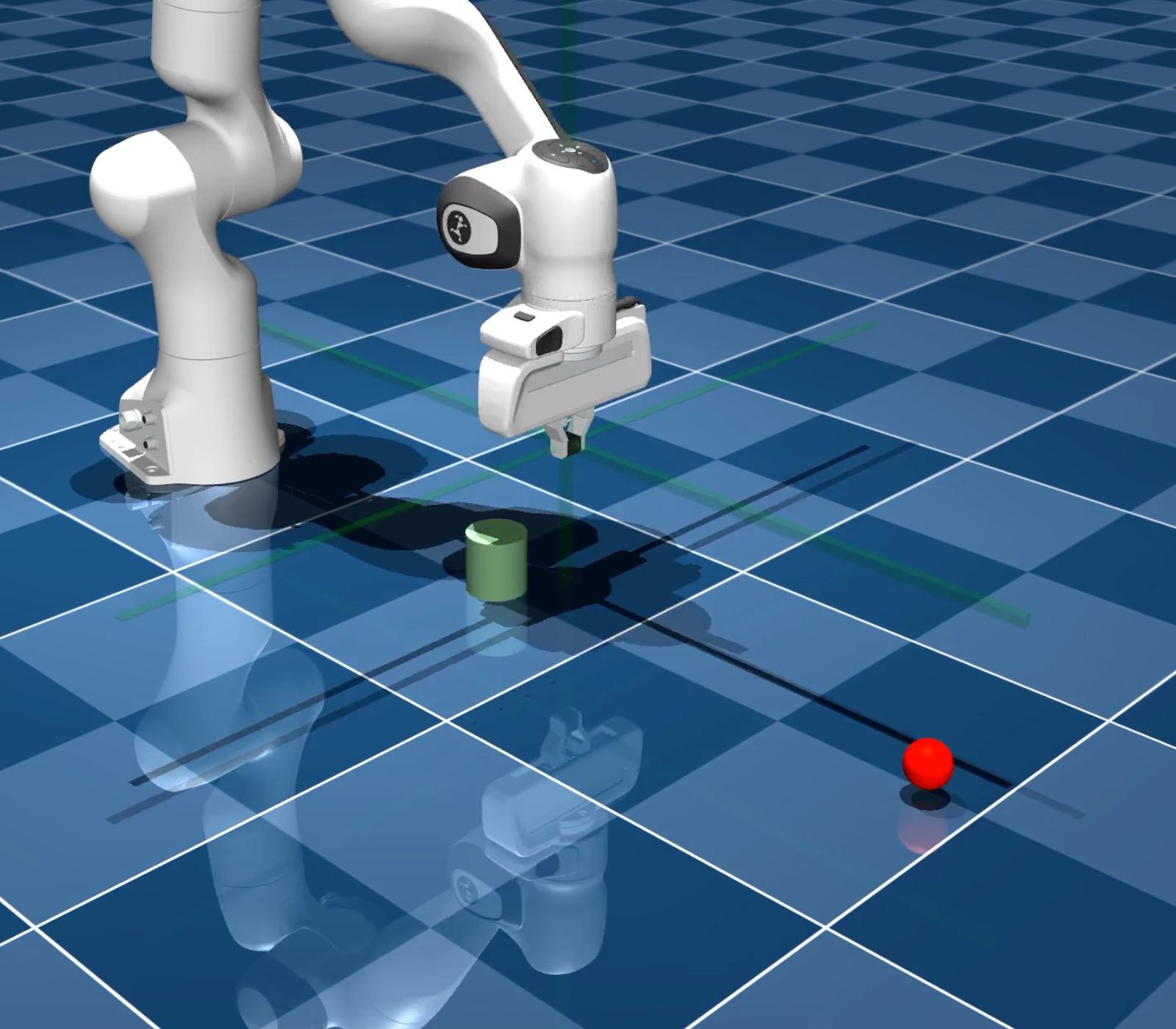}\hspace{0.06cm}
		\includegraphics[scale=0.05]{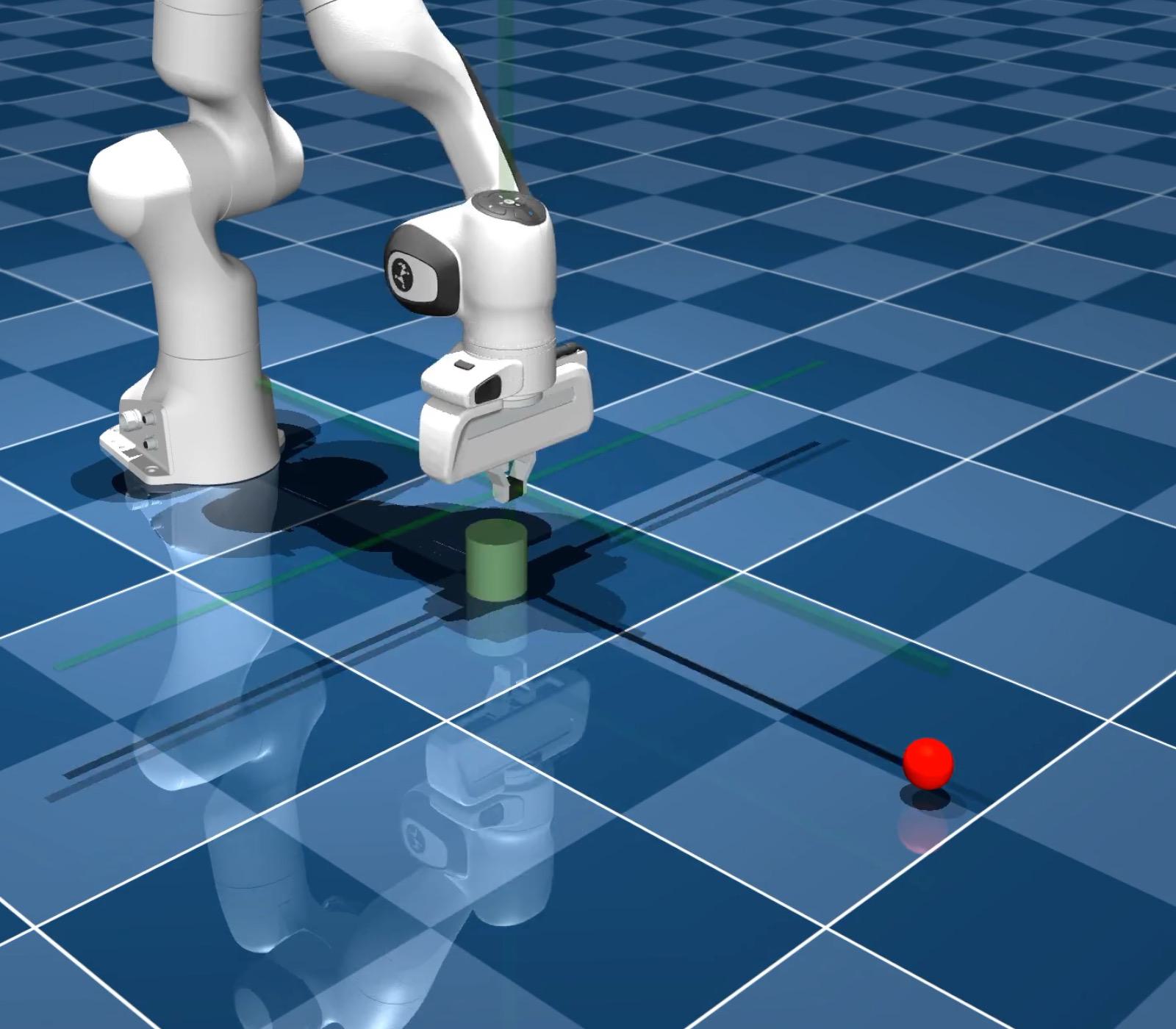}\hspace{0.06cm}
		\includegraphics[scale=0.05]{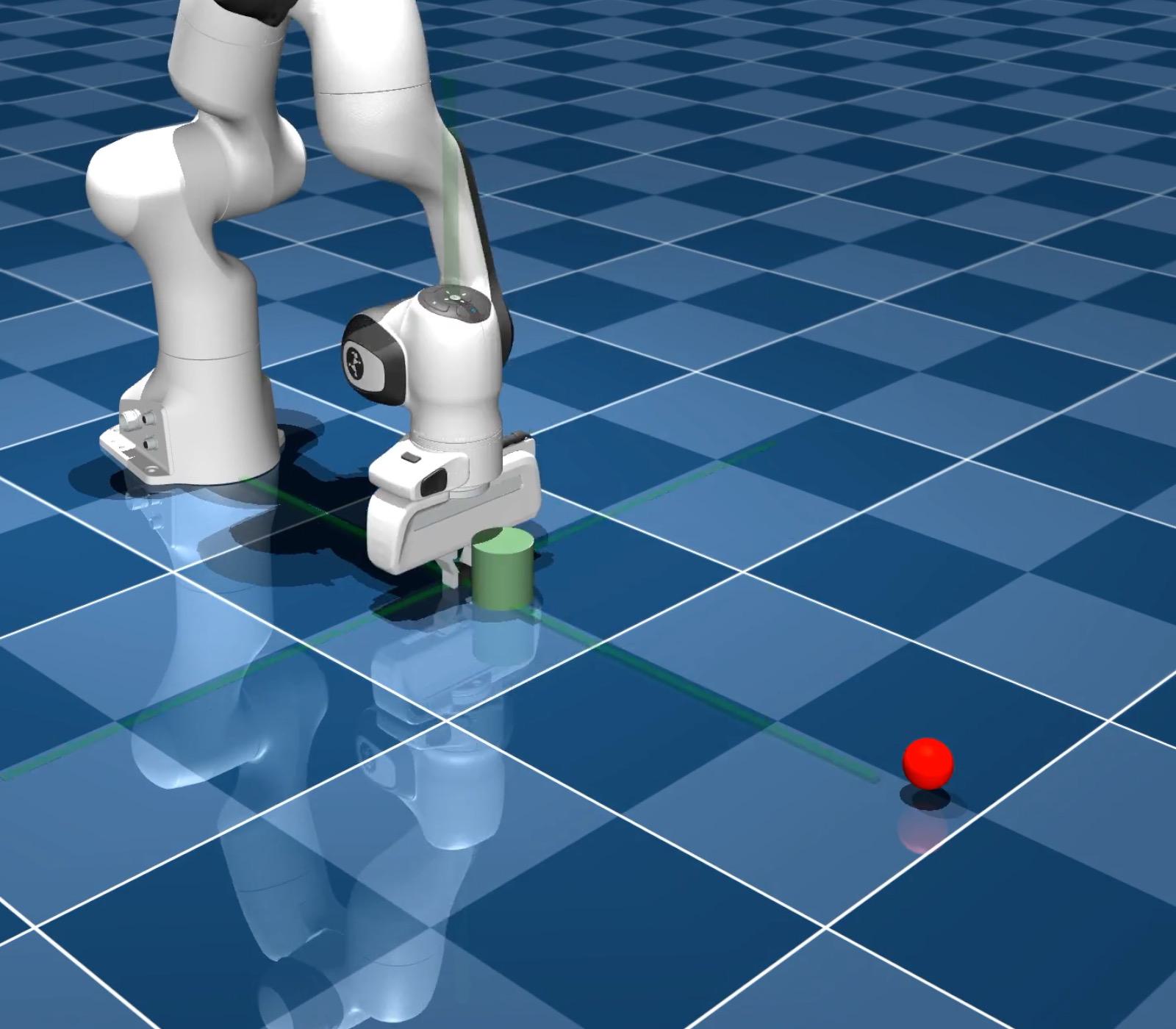}\hspace{0.06cm}
		\includegraphics[scale=0.05]{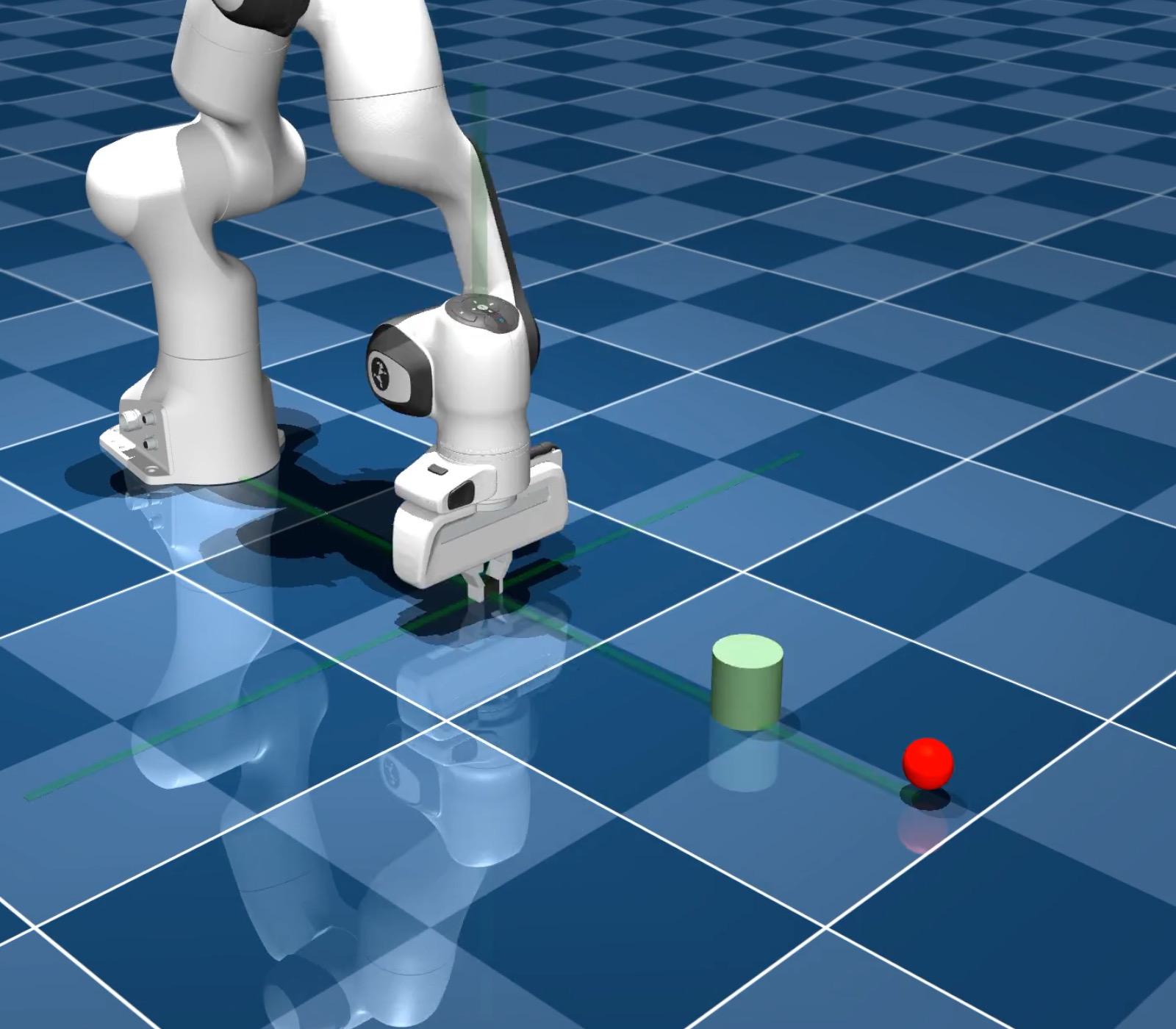}}

	\vspace{0.1cm}
	\captionsetup[subfigure]{skip=4pt,slc=off,margin={212pt, 0pt},labelfont=normalfont}
	\subcaptionbox{FrankaPickAndPlace\label{pick_seq}}[\textwidth][c]{
		\includegraphics[scale=0.05]{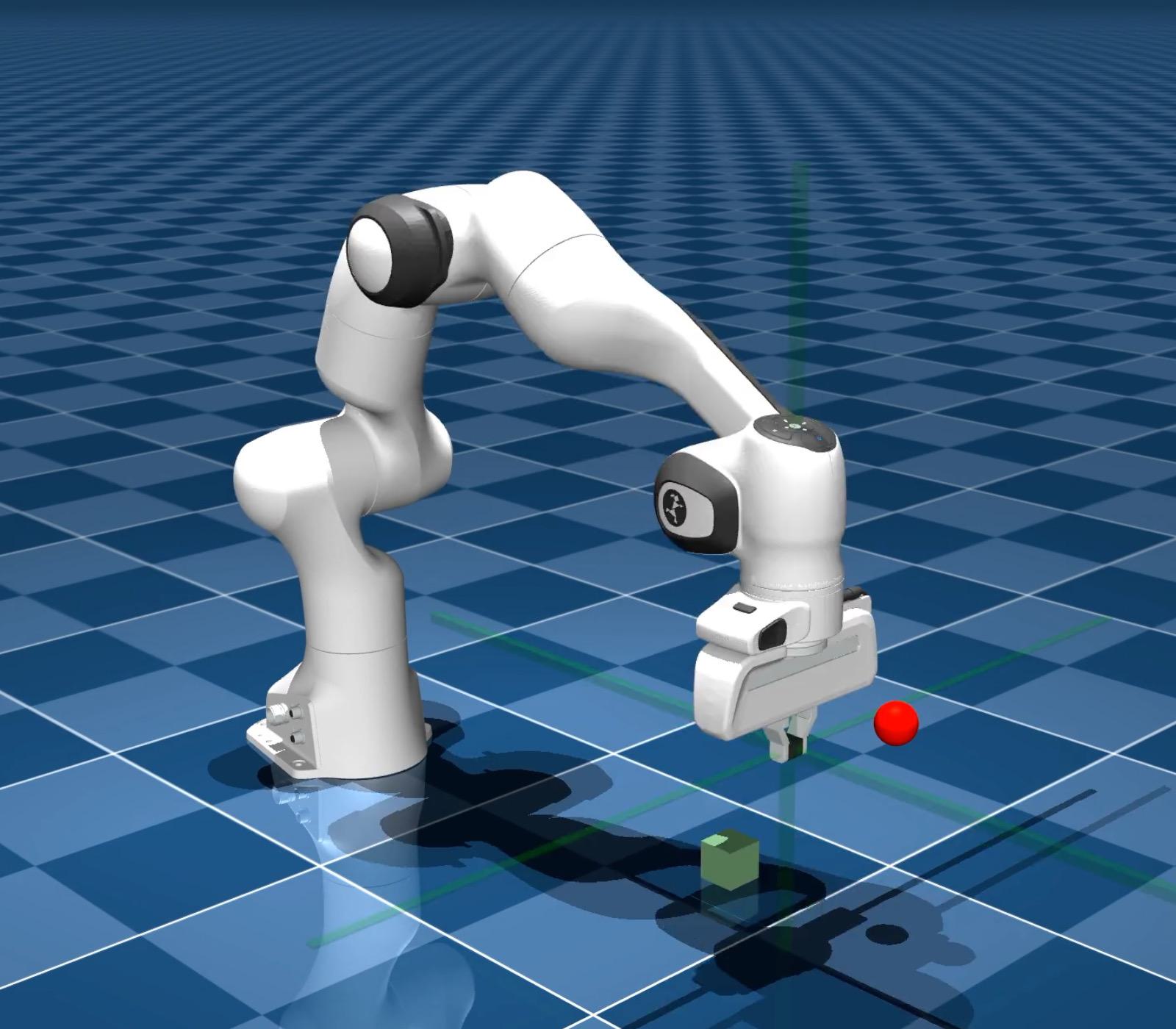}\hspace{0.06cm}
		\includegraphics[scale=0.05]{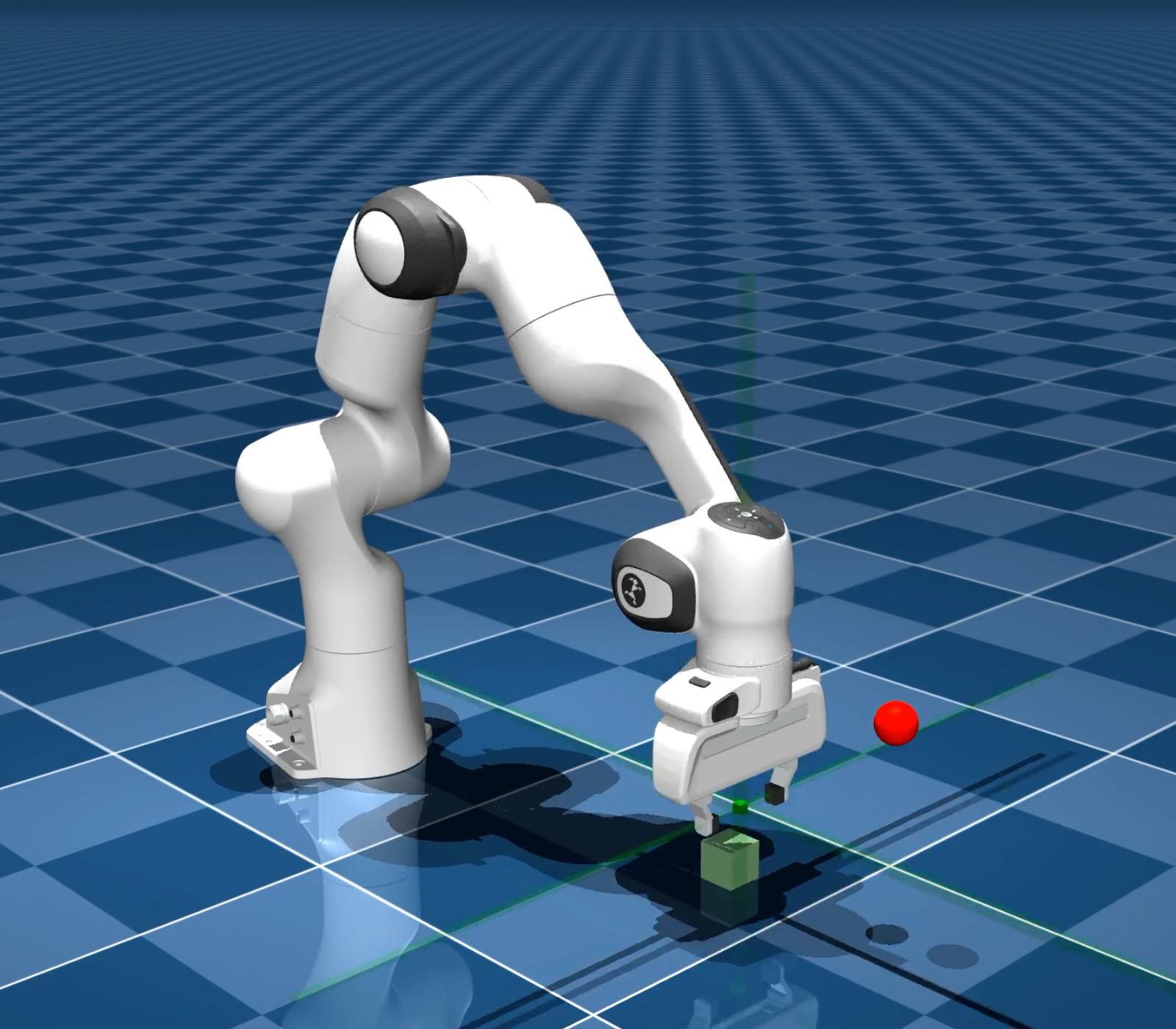}\hspace{0.06cm}
		\includegraphics[scale=0.05]{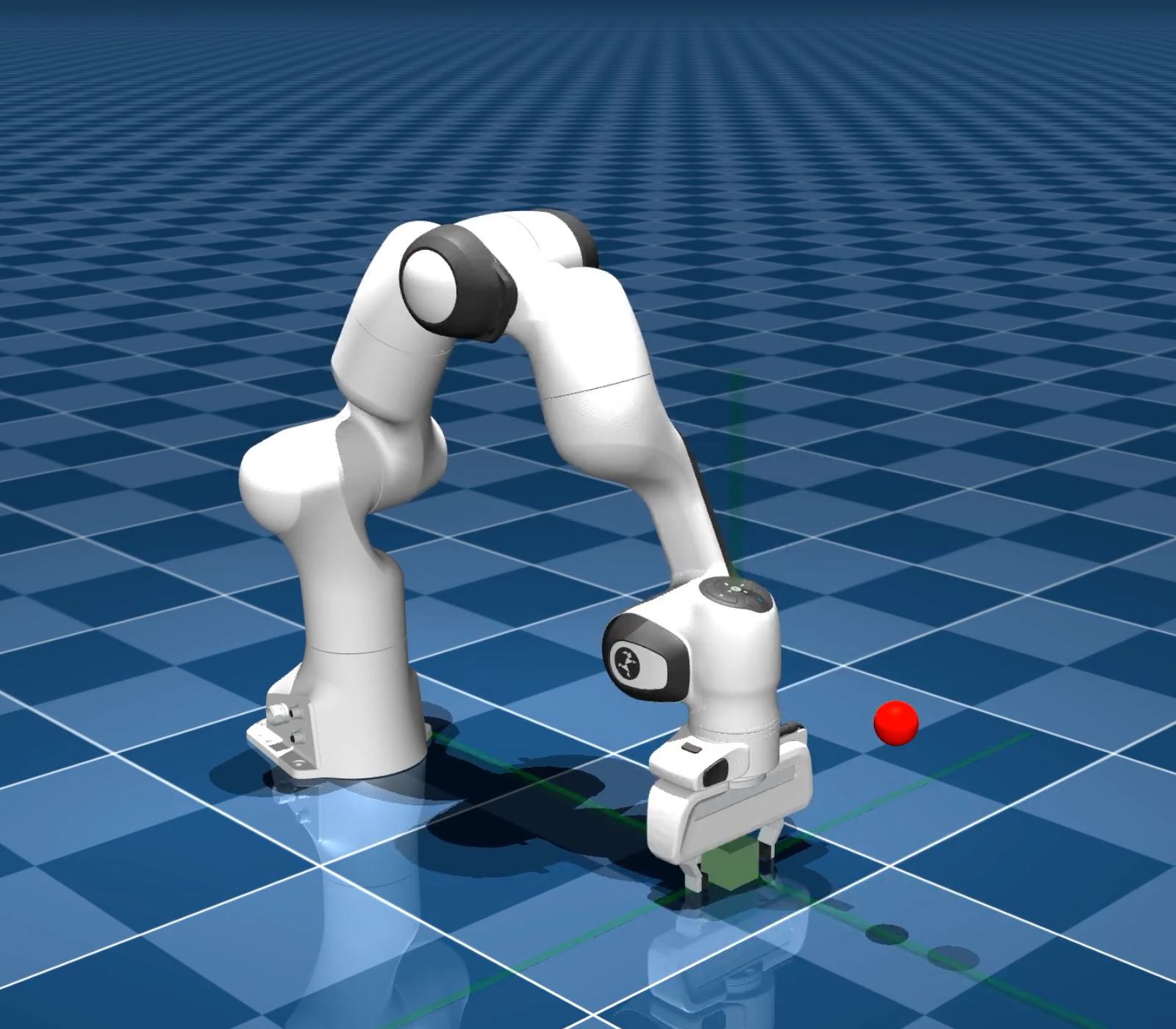}\hspace{0.06cm}
		\includegraphics[scale=0.05]{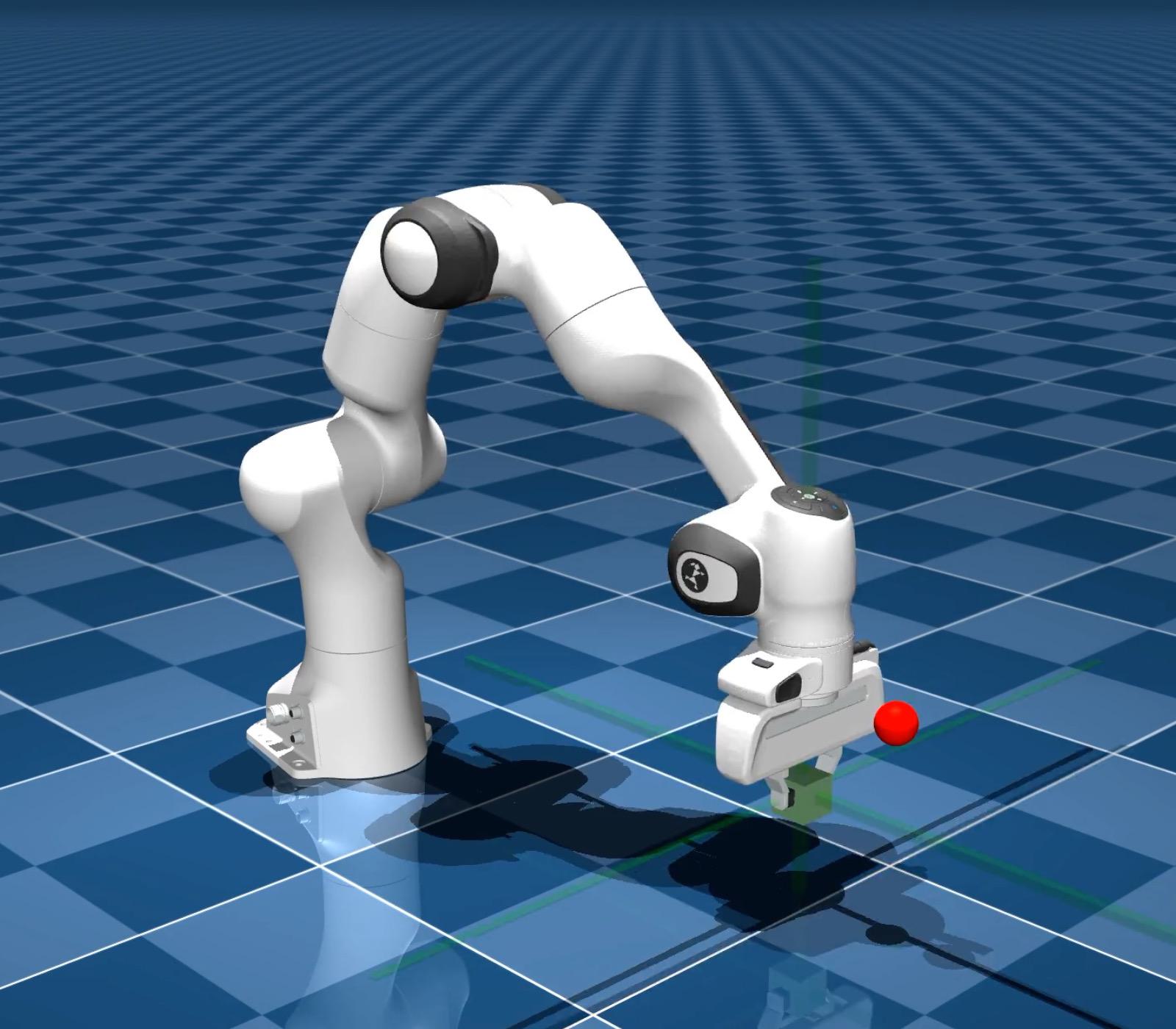}\hspace{0.06cm}
		\includegraphics[scale=0.05]{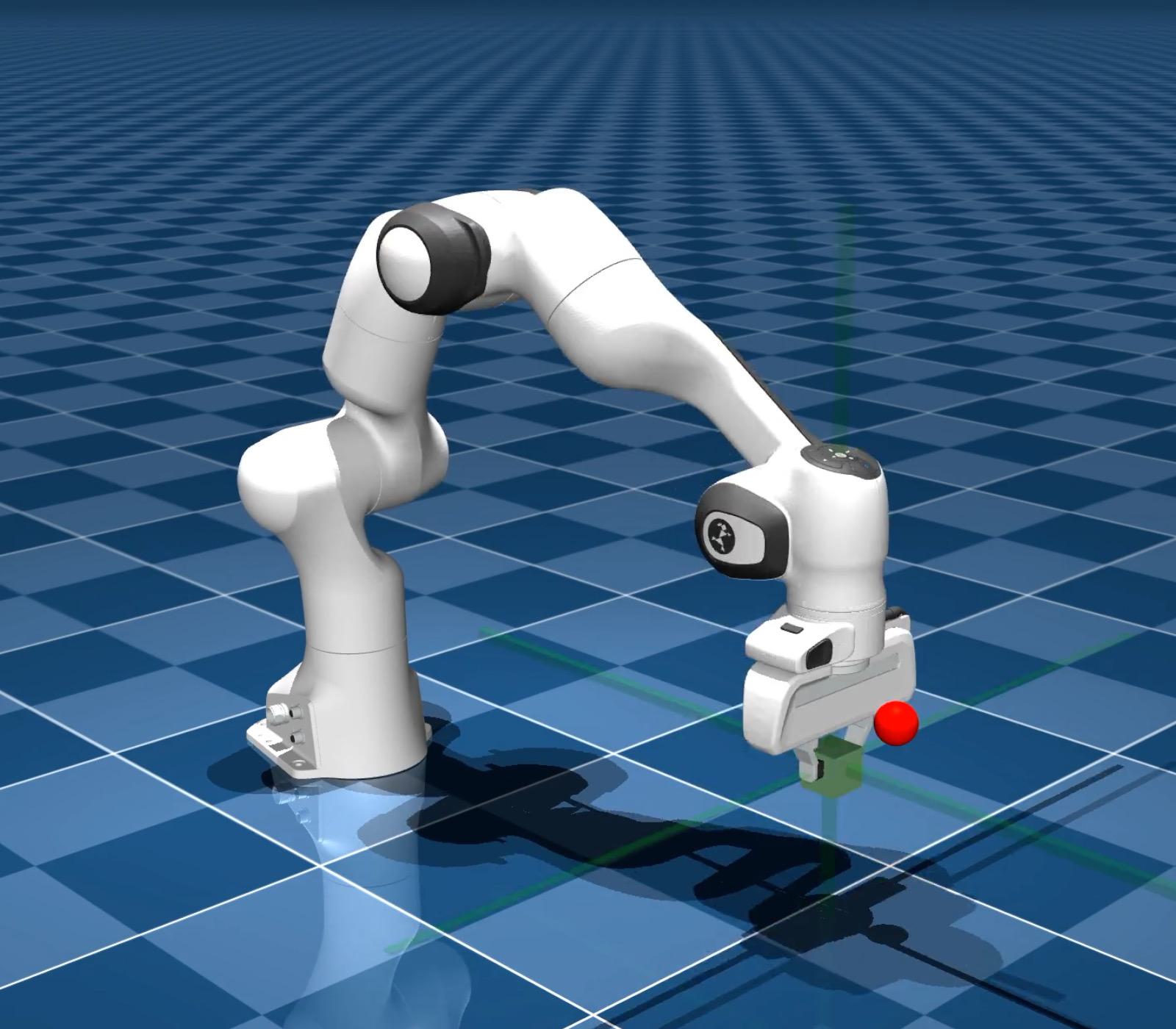}}
	\caption{The execution processes of different tasks via the trained policies.}
	\vspace{-1pt}
	\label{results_seq}
\end{figure*}

This section verifies the feasibility of benchmark environments, i.e., whether the RL algorithm can accomplish these tasks. Thus, three representative off-policy RL algorithms, DDPG, SAC, and TQC, are selected for training. From DDPG to TQC, the main improvements are dedicated to enhancing the performance of continuous policy to alleviate the overestimation bias. Unlike DDPG and SAC, TQC models the distribution over the critic value instead of the Q-function, from which several topmost atoms are truncated. The evaluation results in \cite{kuznetsov2020controlling} indicate that TQC can obtain higher episode returns compared to SAC on MuJoCo environments. The performance of different algorithms when acting on our benchmark environments is also verified in this section. The implementation of the above algorithms is based on Stable-Baselines3\cite{stable-baselines3}.

At the beginning of each episode, the arm is reset to the initial configuration, and the positions of target and goal are randomized within the specified area. For the pick-and-place task, the trick of the target position being randomized on the floor facilitates convergence, and the corresponding random probability can be adjusted. The episode length is 50 by default. Truncated and terminated signals are detected after each step. Termination can end the current episode in advance without wandering up to the maximum step limit. The transitions are saved in the replay buffer with Hindsight Experience Replay (HER) \cite{andrychowicz2017hindsight}. The policy is evaluated via 50 episodes at an interval of 2,000 training steps. Once the object reaches the desired goal point within the episode length, this episode is labeled as successful. Thus, the success rate will serve as the primary evaluation metric. The detailed hyperparameters are available in Table~\ref{hyper}. Actor and critic share the same structure of multilayer perceptron (MLP), and the network size indicated in Table~\ref{hyper} refers to the hidden layers.

Each algorithm is trained over three random seeds, and all training processes are deployed on a workstation with an NVIDIA RTX 3070Ti GPU and an i7-12700K CPU. Fig.~\ref{evaluation} depicts the evaluation curves with the sparse reward for all tasks. DDPG performs the worst across all tasks and its success rate is less than 0.2 in the \texttt{FrankaPush} task. Fig.~\ref{push} typically reveals the disadvantage of DDPG that overestimation bias causes cumulative errors and thus leads to suboptimal strategies and divergent behaviors \cite{fujimoto2018addressing}. Then, clipped double-Q learning is proposed in \cite{fujimoto2018addressing} to reduce the overestimation bias effectively and is incorporated in SAC. The evaluation results in Fig.~\ref{evaluation} also demonstrate that SAC performs better than DDPG across all tasks. Meanwhile, TQC performs the best and achieves a success rate close to 1 after training, which is consistent with the evaluation scores in \cite{kuznetsov2020controlling}. 

For \texttt{FrankaSlide}, the target will be randomly distributed to the unreachable area for the Franka arm, which makes it hard to consistently adjust the object position. Thus, the training takes 1 million steps to improve the agent's performance. As shown in Fig.~\ref{slide}, SAC and TQC stabilize around the $80\%$ success rate after training and rarely increase further, which is similar to the distributions in Fetch and panda-gym. For \texttt{FrankaPush}, SAC and TQC reach a success rate of more than $80\%$ in around $2\times10^5$ steps. Finally, each trained model is tested over 20 random episodes, and the corresponding success rate of each algorithm is counted in Table~\ref{results}. Fig.~\ref{results_seq} illustrates the deterministic episode for each task with the trained TQC model, where the Franka arm will drive the object to reach the goal position from the left to right columns.

\section{CONCLUSION AND FUTURE WORK}\label{cons}
Three open-source environments corresponding to three manipulation tasks are proposed in this paper, where each environment follows the Multi-Goal Reinforcement Learning framework. These benchmark environments are built with the Franka model in MuJoCo Menagerie and powered by MuJoCo. Note that each environment is defined in a clean way in which unnecessary dependencies and modules are omitted. Three representative off-policy algorithms with Hindsight Experience Replay are implemented to validate the feasibility of each environment. To the best of our knowledge, there are a few benchmark environments built on top of MuJoCo Menagerie. We expect that our open-source environments can gain some attention to create more benchmark environments with MuJoCo Menagerie and thus lower barriers for researchers to test algorithms.

However, there are some limitations in our work. More diverse and realistic controllers, like the impedance controller, should be adopted to minimize the sim2real gap. Moreover, a smaller timestep has not been tested in MuJoCo considering the computing resources that we have. Future work will further dive into these limitations to be as faithful as possible to the real environment and offer more test results. Extra tasks and manipulators can also be considered to be added to the current work.
\section{ACKNOWLEDGMENT}
This work was supported by the National Natural Science Foundation of China [grant numbers: 92148203 and T2388101].

\end{document}